\documentclass{article}

\usepackage[preprint]{neurips_2026}

\usepackage[utf8]{inputenc}
\usepackage[T1]{fontenc}
\usepackage{hyperref}
\usepackage{url}
\usepackage{booktabs}
\usepackage{amsmath}
\usepackage{amssymb}
\usepackage{graphicx}
\usepackage[dvipsnames]{xcolor}
\usepackage{colortbl}
\usepackage{caption}
\usepackage{subcaption}
\usepackage{algorithm}
\usepackage{algpseudocode}

\usepackage{xargs}                      
\usepackage[colorinlistoftodos,prependcaption,textsize=tiny]{todonotes}

\newcommand{\codeurl}{\url{https://github.com/armanbolatov/diffusion-baselines}}

\newcommand{\textEightNSig}{3}
\newcommand{\textEightNTied}{3}
\newcommand{\celebaTau}{-0.14}
\newcommand{\celebaMseSpread}{1.8}

\newcommand{\qmNineTauFcd}{-0.43}
\newcommand{\qmNineFcdBest}{Schedule-Free}

\newcommand{\crossTauCelebaSixtyFourTextEight}{+0.71}
\newcommand{\crossTauLMOneBQMNine}{-0.14}

\title{Optimizers for Diffusion Models: A Controlled Benchmark}

\author{%
  Arman Bolatov\textsuperscript{1} \quad
  Egor Shulgin\textsuperscript{2} \quad
  David Li\textsuperscript{3} \quad
  Abduragim Shtanchaev\textsuperscript{3} \\
  \bfseries
  Sebastian U.\ Stich\textsuperscript{1} \quad
  Maxim Panov\textsuperscript{3} \quad
  Eric Moulines\textsuperscript{3, 4} \quad
  Peter Richt\'arik\textsuperscript{2} \\
  \bfseries
  Martin Tak\'a\v{c}\textsuperscript{3} \\[1.5ex]
  \textsuperscript{1}CISPA Helmholtz Center for Information Security \quad
  \textsuperscript{2}KAUST \quad
  \textsuperscript{3}MBZUAI \quad
  \textsuperscript{4}EPITA%
}

\begin{document}
\maketitle

\begin{abstract}
Discrete diffusion models now match autoregressive language models on several benchmarks, while the question of how best to train them has received far less attention: the optimizer is inherited from one paper to the next and never compared. New optimizers, meanwhile, are validated almost exclusively on autoregressive pretraining, a different objective on a different loss surface. We present a controlled optimizer benchmark across four diffusion formulations, to our knowledge the first for discrete diffusion: seven optimizers (AdamW, Lion, Muon, SOAP, MARS, MARS-M, Schedule-Free) on masked diffusion (text8), uniform diffusion (QM9, and LM1B through the Gaussian duality) and Gaussian diffusion on images (CelebA-64), each on a task with published reference values. Every optimizer receives the same search protocol, and every winner is retrained at the full budget with three seeds. AdamW is a strong default but not always the right choice: it is beaten by a resolved margin on two of the four tasks, and the winner changes with the formulation, so the optimizer deserves the same care as the rest of the training recipe. Notably, methods validated on autoregressive language model pretraining transfer well: Muon, MARS-M and SOAP each beat the tuned AdamW on at least one diffusion formulation. The benchmark, all runs and every figure are reproducible end to end from the released code at \codeurl.
\end{abstract}

\section{Introduction}

New optimizers are usually validated on autoregressive language model
pretraining or on supervised vision and speech. Muon \citep{jordan2024muon},
SOAP \citep{vyas2025soap} and MARS \citep{yuan2025mars} report their gains on
autoregressive language models, Schedule-Free
\citep{defazio2024schedulefree} won the self-tuning track of the AlgoPerf
competition \citep{dahl2023algoperf,kasimbeg2025algoperf}, whose workloads
include no diffusion model, and
the benchmarks that compare optimizers head to head
\citep{schmidt2021crowded,semenov2025benchmarking} likewise contain no
diffusion task. Diffusion models \citep{sohldickstein2015diffusion,ho2020ddpm,song2021sde} are
trained with a different objective, on a different loss surface, and often with
a different backbone, a U-Net \citep{ronneberger2015unet} or a diffusion
transformer \citep{peebles2023dit} rather than a causal decoder. Whether the
reported gains carry over is open.

Diffusion offers language models capabilities autoregression lacks: a
diffusion language model denoises the whole sequence at every step with
bidirectional context, so it can decode many tokens in parallel, generate in
any order, fill in the middle of a text, and revise tokens it has already
produced \citep{gong2025diffullama,schiff2025guidance}. Trained from scratch,
such models reach the level of strong autoregressive models at the 8B scale
\citep{nie2025llada}, and adapted from pretrained autoregressive models they
match their sources on most benchmarks \citep{gong2025diffullama}.

The question matters for discrete diffusion in particular. MDLM, UDLM
and DUO \citep{sahoo2024mdlm,schiff2025guidance,sahoo2025duo} all train with
AdamW, with the same hyperparameters passed from one paper to the next rather
than compared against alternatives. A practitioner training a diffusion
language model today has no evidence about whether a different optimizer would
help.

We run a controlled benchmark of seven optimizers across four diffusion
formulations: masked diffusion on text8, uniform diffusion on QM9, uniform
diffusion trained through the Gaussian duality (DUO) on LM1B, and Gaussian
image diffusion (DDPM) on CelebA-64. Two of the discrete tasks are natural
language and one is the SMILES molecular language, and every formulation runs
on a standard task with published reference values
(Section~\ref{sec:formulations}). Our contributions are:

\begin{enumerate}
\item To our knowledge, the first controlled optimizer benchmark for discrete
diffusion models, run alongside Gaussian image diffusion: seven optimizers,
AdamW, Lion, Muon, SOAP, MARS,
MARS-M and Schedule-Free AdamW, under one search protocol, with every winner
retrained at the full budget on three seeds (protocol in
Section~\ref{sec:protocol} and Appendix~\ref{app:grids}).
\item A reproducible benchmarking codebase that runs the search, the retraining
and every figure and table in this paper end to end, released at \codeurl.
\end{enumerate}

\section{Related work}

\paragraph{Benchmarking optimizers.} \citet{schmidt2021crowded} evaluated
fifteen optimizers under a fixed tuning budget and found no single method that
won everywhere, and \citet{choi2019empirical} showed that the ranking itself
turns on the tuning protocol. AlgoPerf \citep{dahl2023algoperf,kasimbeg2025algoperf}
set up a time-to-target protocol across several workloads but contains no
diffusion task.
\citet{semenov2025benchmarking} compare eleven optimizers on language model
pretraining from 124M to 720M parameters and find AdEMAMix \citep{pagliardini2025ademamix}, MARS and
D-Muon \citep{liu2025moonlight} ahead of AdamW. \citet{wen2025fantastic} tune ten optimizers with per-optimizer
coordinate descent and conclude that reported speedups over AdamW are smaller
than claimed and shrink with scale. All of this work is on discriminative or
autoregressive models.

\paragraph{Optimizers for diffusion.} \citet{schaipp2025diffusion} benchmarks six
optimizers on a denoising task over Navier-Stokes trajectories and reports Muon
and SOAP reaching 18\% lower final loss than AdamW. That paper also observes
that Schedule-Free reaches a good loss while producing worse samples, which is
the effect we measure on CelebA-64 with a full seven-optimizer ranking. On the
theory side, \citet{tyurin2026reversefisher} show that the choice of score
matching objective changes whether gradient descent converges globally, even on
Gaussian mixtures, a reminder that the denoising and score-matching objectives
diffusion models train need not share the optimization landscape of the
autoregressive cross-entropy.
\citet{chen2026cmuon} show that plain Muon plateaus late in diffusion
transformer training because the fused QKV and MLP tensors of a
transformer \citep{vaswani2017attention} couple subspaces during
orthogonalization, and propose a chunked variant. Our backbone uses a fused QKV
projection and our Muon is the plain, unchunked port
(Appendix~\ref{app:optimizers}), so any such plateau would act on our text8 and
QM9 runs, where Muon nevertheless ties with or beats AdamW.

\paragraph{Diffusion language models.} The formulations we train are masked
diffusion \citep{sahoo2024mdlm,austin2021d3pm,lou2024sedd,shi2024md4} and
uniform diffusion, trained either directly \citep{schiff2025guidance},
building on the categorical flows of \citet{hoogeboom2021argmax}, or through
the Gaussian duality of \citet{sahoo2025duo}; Section~\ref{sec:formulations}
states their objectives. Masked diffusion also extends beyond natural
language, to molecule generation among others \citep{lee2025genmol}. A
separate line obtains diffusion language models at scale, by converting
pretrained autoregressive models \citep{gong2025diffullama,ye2025dream} or by
training from scratch \citep{nie2025llada}. \citet{vonrutte2025scaling} study
how the formulations scale, tuning batch size and learning rate carefully, but
hold the optimizer fixed. We take the complementary view and vary the
optimizer while holding the formulation fixed.

\section{The optimizers}
\label{sec:optimizers}

All seven methods consume the same stochastic gradient $g_t = \nabla f(\theta_{t-1}, \xi_t)$
and differ in how they turn it into a step on the parameters $\theta$. Below,
$\eta$ is the learning rate, $\lambda$ the decoupled weight decay, $m_t$ and
$v_t$ are exponential moving averages with rates $\beta_1, \beta_2$, and all
operations between tensors of the same shape are element-wise; we omit bias
corrections for clarity. Steepest descent under a norm comes in two forms
\citep{crawshaw2025exploration}: constrained, where the step is a linear
minimization over a norm ball and has fixed size, and regularized, where the
step scales with the gradient's dual norm. \citet{veprikov2025preconditioned}
extend this view with preconditioned norms; we use it to place the seven.

\textbf{AdamW} \citep{kingma2015adam,loshchilov2019adamw} preconditions a
momentum step by a running second-moment estimate
\citep{duchi2011adagrad,tieleman2012rmsprop}:
\begin{equation}
m_t = \beta_1 m_{t-1} + (1-\beta_1) g_t, \qquad
v_t = \beta_2 v_{t-1} + (1-\beta_2) g_t^2, \qquad
\theta_t = (1 - \eta\lambda)\,\theta_{t-1} - \eta\, \frac{m_t}{\sqrt{v_t} + \epsilon}.
\end{equation}
The preconditioner changes every iteration, so this is regularized steepest
descent in a moving metric rather than under a fixed norm
\citep{schaipp2024momo,crawshaw2025exploration}. The weight decay
\citep{krogh1991weight} is decoupled: $\theta$ is shrunk outside the adaptive
step instead of adding $\lambda\theta$ to $g_t$ where the preconditioner would
rescale it.

\textbf{Lion} \citep{chen2023lion}, found by symbolic program search over
update rules, signs a momentum, but with two interpolation rates, one for the
update and one for the stored state:
\begin{equation}
\theta_t = \theta_{t-1} - \eta \big( \operatorname{sign}(\beta_1 m_{t-1} + (1-\beta_1) g_t) + \lambda\,\theta_{t-1} \big),
\qquad
m_t = \beta_2 m_{t-1} + (1-\beta_2) g_t,
\end{equation}
with $\beta_2 > \beta_1$ (defaults $0.99$ and $0.9$; both are tuned here),
which weights the current gradient about tenfold against a plain moving
average. With decoupled decay, Lion provably minimizes the loss subject to
$\lVert\theta\rVert_\infty \le 1/\lambda$ \citep{chen2024lionk}.

\textbf{Muon} \citep{jordan2024muon} is the one constrained method here, the
linear minimizer over a spectral-norm ball
\citep{bernstein2024norm,pethick2025scion}: for a hidden weight matrix with
momentum $M_t = \mu M_{t-1} + G_t$,
\begin{equation}
\theta_t = (1-\eta\lambda)\,\theta_{t-1} - \eta'\, \mathrm{NS}_5(M_t),
\qquad
\mathrm{NS}_5(M) \approx U V^{\!\top} \text{ for } M = U \Sigma V^{\!\top},
\end{equation}
where $\mathrm{NS}_5$ is a five-step Newton--Schulz iteration and
$\eta' = 0.2\,\eta\sqrt{\max(m,n)}$ matches the update scale to AdamW's
\citep{liu2025moonlight}. Embeddings, the output head and vectors take an
AdamW step with their own learning rate.

\textbf{SOAP} \citep{vyas2025soap} runs the Adam update inside the eigenbasis
$(Q_L, Q_R)$ of Shampoo's \citep{gupta2018shampoo} two Kronecker factors, the
curvature factorization introduced by K-FAC \citep{martens2015kfac}:
\begin{equation}
\theta_t = \theta_{t-1}
- \eta\, Q_L\!\left( \frac{\tilde{M}_t}{\sqrt{\tilde{V}_t} + \epsilon} \right)\! Q_R^{\!\top}
- \eta\lambda\,\theta_{t-1},
\qquad \tilde{M}_t = Q_L^{\!\top} M_t Q_R,
\end{equation}
with the basis refreshed every \texttt{precondition\_frequency} steps. Fixing
$Q_L = Q_R = I$ recovers Adam; the refresh is where its extra cost lives.

\textbf{MARS} \citep{yuan2025mars} changes the gradient rather than the
geometry, scaling a STORM-style \citep{cutkosky2019storm} variance-reduction
correction by \texttt{gamma}:
\begin{equation}
\label{eq:mars}
c_t = g_t + \gamma\,\frac{\beta_1}{1-\beta_1}\,(g_t - g_{t-1}),
\end{equation}
then takes the AdamW step with $c_t$ in place of $g_t$ in both moments;
$\gamma = 0$ recovers AdamW, and scaling the correction well below one is what
makes it work where earlier variance reduction did not help deep networks
\citep{defazio2019ineffectiveness}. \textbf{MARS-M} \citep{liu2025marsm} feeds the same
$c_t$ into Muon's step, so the MARS--AdamW gap isolates the correction
exactly, and the MARS-M--Muon gap approximately (their momentum damping
differs, Appendix~\ref{app:optimizers}).

\textbf{Schedule-Free AdamW} \citep{defazio2024schedulefree} keeps Adam's
preconditioner and replaces momentum with an interpolation between a base
iterate $z_t$, where the step is taken, and its running average $x_t$, which
is what gets evaluated \citep{polyak1992averaging}:
\begin{equation}
y_t = (1-\beta_1) z_t + \beta_1 x_t,
z_{t+1} = z_t - \eta_t \frac{\nabla f(y_t, \xi_t)}{\sqrt{v_t}+\epsilon} - \eta_t \lambda y_t,
x_{t+1} = (1 - c_{t+1})\, x_t + c_{t+1} z_{t+1},
\end{equation}
with weights $c_{t+1} \propto \eta_t^2$. The average stands in for a decay
schedule, so the learning rate stays flat and the run needs no fixed end
point. Full pseudocode for all seven is in
Appendix~\ref{app:pseudocode}.

\section{Setup}

\subsection{Datasets}

We use four datasets, one per diffusion formulation (Table~\ref{tab:tasks},
Appendix~\ref{app:arch});
Appendix~\ref{app:data} gives the preprocessing in full. \textbf{text8}
\citep{mahoney2011text8} is 100M characters of cleaned Wikipedia over a
27-symbol alphabet, split 90M/5M/5M, trained on windows of 256. \textbf{QM9}
\citep{ramakrishnan2014qm9} is 134k small molecules as canonical SMILES strings
\citep{weininger1988smiles}, tokenized to a 40-symbol vocabulary and padded to
length 32. \textbf{LM1B} \citep{chelba2013lm1b} is sentence-level news text,
tokenized with the \texttt{bert-base-uncased} WordPiece vocabulary
\citep{devlin2019bert,wolf2020transformers} at length 128. \textbf{CelebA}
\citep{liu2015celeba} is 203k aligned face photographs, center-cropped to 140
pixels and resized to $64\times64$; its training set provides the FID reference
statistics \citep{heusel2017fid}.

\subsection{Diffusion formulations and models}
\label{sec:formulations}

\paragraph{Why one formulation per task.} Each discrete formulation is
evaluated on the task its original paper positions it on: text8 has been the
standard testbed for discrete diffusion since D3PM
\citep{austin2021d3pm,lou2024sedd,shi2024md4}, QM9 is the flagship result of
UDLM, and LM1B is the headline task of DUO; Gaussian diffusion runs on CelebA
at $64\times64$, one of the benchmarks of the DDIM paper \citep{song2021ddim}.
This gives every pair published reference values (Appendix~\ref{app:budgets}) and
keeps the question clean: we rank optimizers within a formulation, not
formulations against each other; whether the ranking is consistent across the
setups is measured in Section~\ref{sec:transfer}.

All three discrete formulations train a continuous-time bound on the negative
log-likelihood \citep{kingma2021vdm,campbell2022continuous} of the form
\begin{equation}
\label{eq:nelbo}
\mathcal{L}(\theta)
= \mathbb{E}_{t \sim \mathcal{U}[0,1]}\;
  \mathbb{E}_{z_t \sim q_t(\cdot \mid x)}
  \big[\, w(t)\; \ell\big(x, \, p_\theta(\cdot \mid z_t, t)\big) \big],
\end{equation}
where $x$ is the clean token sequence, $t \in [0,1]$ the diffusion time,
$z_t$ the corrupted sequence drawn from the corruption kernel
$q_t(\cdot \mid x)$, $p_\theta$ the denoising model, $w(t)$ a time-dependent
weight, and $\ell$ a per-token reconstruction term. The three formulations
differ only in $q_t$, $w$ and $\ell$; we report the bound in each task's
customary unit, bits per character on text8 \citep{mikolov2012subword} and
per-token perplexity on QM9 and LM1B \citep{jelinek1977perplexity}.

\textbf{text8 with MDLM} \citep{sahoo2024mdlm}. Masked diffusion replaces each
token with a mask symbol $\mathrm{m}$ independently with probability
$1-\alpha_t$; the model fills masked positions given the rest. With the linear
schedule $\alpha_t = 1-t$, Eq.~\eqref{eq:nelbo} becomes a weighted
cross-entropy on the masked positions,
$w(t) = 1/t$ and
$\ell = -\sum_{i:\, z_t^i = \mathrm{m}} \log p_\theta(x^i \mid z_t)$.

\textbf{QM9 with UDLM} \citep{schiff2025guidance}. Uniform diffusion
resamples a corrupted token uniformly from the vocabulary, so, unlike masking,
a corrupted token carries no flag and the model must also decide which
positions to revise. Here $\ell$ is the evidence bound of
\citet{schiff2025guidance}, a cross-entropy between the analytic posterior and
the model posterior with $w(t) = -\alpha_t'/(N\alpha_t)$ for vocabulary size
$N$.

\textbf{LM1B with DUO} \citep{sahoo2025duo}. DUO trains the same uniform
diffusion bound, obtained through the diffusion duality: uniform diffusion is
the image of a Gaussian diffusion under an argmax. The model trains against
the underlying Gaussian process for the first half of training, then switches
to the discrete objective; we follow the original schedule.

\textbf{CelebA-64 with DDPM} \citep{ho2020ddpm}. Gaussian
diffusion adds pixel noise over 1000 steps on a linear $\beta$ schedule and the
model predicts the noise, with the simple loss
$\mathbb{E}_{t,\epsilon}\lVert \epsilon - \epsilon_\theta(x_t, t)\rVert^2$.
Because this loss is a poor guide to image quality, we also report FID
\citep{heusel2017fid}.

\paragraph{Architectures.} The three text and molecule tasks share one
backbone: a bidirectional pre-norm transformer \citep{vaswani2017attention}
with learned position embeddings, conditioned on the diffusion time through
zero-initialized adaptive layer norm as in DiT \citep{peebles2023dit}. CelebA
uses the DDPM U-Net as implemented in \texttt{diffusers}
\citep{vonplaten2022diffusers}. Sizes are in Table~\ref{tab:tasks} and the full
architecture detail in Appendix~\ref{app:arch}.

All four share one training loop in PyTorch \citep{paszke2019pytorch}: bfloat16
autocast \citep{micikevicius2018mixed}, a warmup-stable-decay schedule
\citep{hu2024minicpm,hagele2024scaling} with an untuned 10\% warmup (the best
warmup length is optimizer- and batch-dependent \citep{riabinin2026warmup},
so one fixed fraction keeps the comparison even but is optimal for none), and
a divergence guard that stops a run whose loss becomes non-finite.
Schedule-Free uses a constant learning rate after the same warmup, since a
decay schedule would defeat its purpose.

\subsection{Tuning protocol}
\label{sec:protocol}

Every optimizer gets the same treatment, since the outcome of an optimizer
comparison depends on how the tuning budget is spent
\citep{choi2019empirical,sivaprasad2020tuning}. The search space is a discrete grid for \texttt{lr}, \texttt{weight\_decay}
and \texttt{grad\_clip}, common to the optimizers up to the per-study
deviations listed in Appendix~\ref{app:grids} \citep{pascanu2013difficulty}, plus the axes specific to each method (Table~\ref{tab:grid}). We use
Optuna \citep{akiba2019optuna} with a multivariate TPE sampler
\citep{bergstra2011tpe,watanabe2023tpe} and Hyperband pruning
\citep{li2018hyperband}, running
100 trials per optimizer on text8 and QM9 and 50 on CelebA-64 and LM1B.
Trials run at 20\% of the full budget, the standard
economy of tuning on a cheaper proxy
\citep{yang2022mutransfer,hu2024minicpm,dahl2023algoperf}; the proxy only
selects configurations, and every number we report comes from full-budget
retraining.

Where a selected value sat on a grid edge we widened the boundary and re-ran
the study; ten selections still sit on an edge (Appendix~\ref{app:grids}),
and three of the four CelebA-64 cases are on axes whose widened grids were
never searched, so the Muon and MARS-M results there come from the narrower
grid the study ran.

The winning configuration is then retrained at the full budget with three
seeds, and we report the final evaluation with the standard deviation across
them. We never report the best point along a trajectory, since that is a
selection over the same evaluation set, and we report the full search budget
rather than the winning configuration alone \citep{dodge2019showyourwork}.
Tuning and benchmarking together used roughly 1400 GPU-hours on NVIDIA A6000,
A100 and RTX PRO 6000 GPUs.

\section{Results}
\label{sec:results}

\begin{table}[t]
\caption{Final metric after full-budget training, mean and standard deviation
over three seeds. Lower is better; bold is the best per task. Cells
are shaded on a diverging scale centered on AdamW, green better and red worse,
the tint deepening with the size of the gap relative to each task's spread;
arrows mark the differences that the paired within-seed test resolves (beyond
twice the standard error). Lion's LM1B entry diverges on two of three seeds
(Section~\ref{sec:lm1b}).}
\label{tab:main}
\centering
\small
\begin{tabular}{lcccc}
\toprule
 & text8 (MDLM) & QM9 (UDLM) & LM1B (DUO) & CelebA-64 (DDPM) \\
 & bpc $\downarrow$ & PPL $\downarrow$ & PPL $\downarrow$ & FID $\downarrow$ \\
\midrule
MARS-M & \cellcolor[HTML]{E0F2E9}\textbf{1.8072 $\pm$ 0.0089} & \cellcolor[HTML]{F7D6D3}2.076 $\pm$ 0.014$^{\downarrow}$ & \cellcolor[HTML]{D3EDE0}62.2 $\pm$ 0.4$^{\uparrow}$ & \cellcolor[HTML]{E2F3EB}\textbf{6.32 $\pm$ 0.62} \\
SOAP & \cellcolor[HTML]{E0F3EA}1.8073 $\pm$ 0.0100 & \cellcolor[HTML]{C3E7D5}\textbf{2.066 $\pm$ 0.013}$^{\uparrow}$ & \cellcolor[HTML]{FCEFEE}69.5 $\pm$ 1.5$^{\downarrow}$ & \cellcolor[HTML]{F4C6C3}10.52 $\pm$ 1.41$^{\downarrow}$ \\
Muon & \cellcolor[HTML]{EDF8F3}1.8126 $\pm$ 0.0040 & \cellcolor[HTML]{D1ECDF}2.069 $\pm$ 0.014$^{\uparrow}$ & \cellcolor[HTML]{D0ECDE}\textbf{61.6 $\pm$ 0.5}$^{\uparrow}$ & \cellcolor[HTML]{F8D8D6}8.58 $\pm$ 0.99$^{\downarrow}$ \\
AdamW & 1.8154 $\pm$ 0.0053 & 2.072 $\pm$ 0.013 & 68.2 $\pm$ 0.8 & 6.88 $\pm$ 0.71 \\
Lion & \cellcolor[HTML]{F5CDC9}1.8509 $\pm$ 0.0064$^{\downarrow}$ & \cellcolor[HTML]{D0ECDE}2.068 $\pm$ 0.014$^{\uparrow}$ & \cellcolor[HTML]{EB968F}diverges & \cellcolor[HTML]{F3BFBB}11.52 $\pm$ 3.37$^{\downarrow}$ \\
MARS & \cellcolor[HTML]{F5CBC8}1.8526 $\pm$ 0.0042$^{\downarrow}$ & \cellcolor[HTML]{F9E2E0}2.074 $\pm$ 0.013$^{\downarrow}$ & \cellcolor[HTML]{E8F6EF}66.6 $\pm$ 0.6$^{\uparrow}$ & \cellcolor[HTML]{F2BCB7}12.01 $\pm$ 1.70$^{\downarrow}$ \\
Schedule-Free & \cellcolor[HTML]{EFA9A3}1.9187 $\pm$ 0.0012$^{\downarrow}$ & \cellcolor[HTML]{F1B3AE}2.085 $\pm$ 0.011$^{\downarrow}$ & \cellcolor[HTML]{EFAEA8}101.5 $\pm$ 5.3$^{\downarrow}$ & \cellcolor[HTML]{EFA9A3}15.33 $\pm$ 1.54$^{\downarrow}$ \\
\bottomrule
\end{tabular}

\end{table}

\subsection{text8: three alternatives tie with AdamW, three lose}

\begin{figure}[t]
\centering
\includegraphics[width=\textwidth]{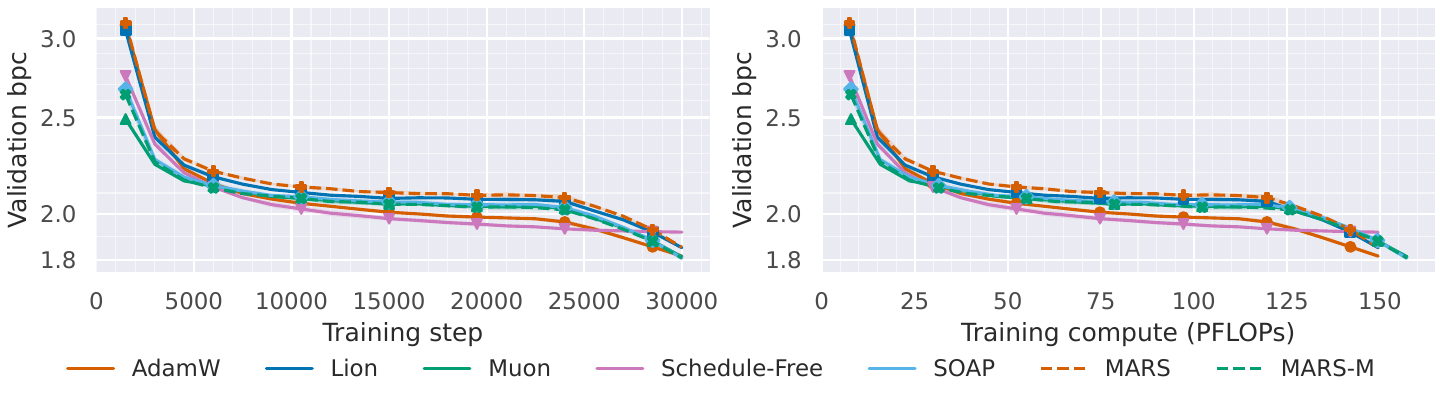}
\caption{text8 validation bits per character on a log scale, mean over three
seeds with a $\pm1$ standard deviation band, against training steps (left) and cumulative
training compute (right). The FLOP counts differ by at most 7\% across
optimizers, so the two panels nearly coincide; per-step wall-clock does not
follow FLOPs (Table~\ref{tab:cost}).}
\label{fig:text8}
\end{figure}

MARS-M, SOAP and Muon finish within 0.009 bits per character of AdamW, and the
paired test does not separate any of them from it (Table~\ref{tab:main});
Lion, MARS and Schedule-Free are worse by 0.036, 0.037 and 0.103, all well
outside seed variation. Of the six alternatives, \textEightNTied{} match AdamW,
\textEightNSig{} lose to it, and none beats it. The geometry-aware methods lead
only over roughly the first 6k steps (Figure~\ref{fig:text8}).

Schedule-Free has the lowest bits per character at every evaluation from 7.5k
through 25.5k steps, but it has no decay phase, and the other six, whose
schedules decay over the final 6k steps, all pass it by the end
(Figure~\ref{fig:text8}). Newer schedule-free variants, such as
Schedule-Free+ \citep{defazio2026schedulefreeplus} and a schedule-free Muon,
would be interesting to test in future work.

\subsection{QM9: three alternatives beat AdamW, by a little}

\begin{figure}[t]
\centering
\includegraphics[width=\textwidth]{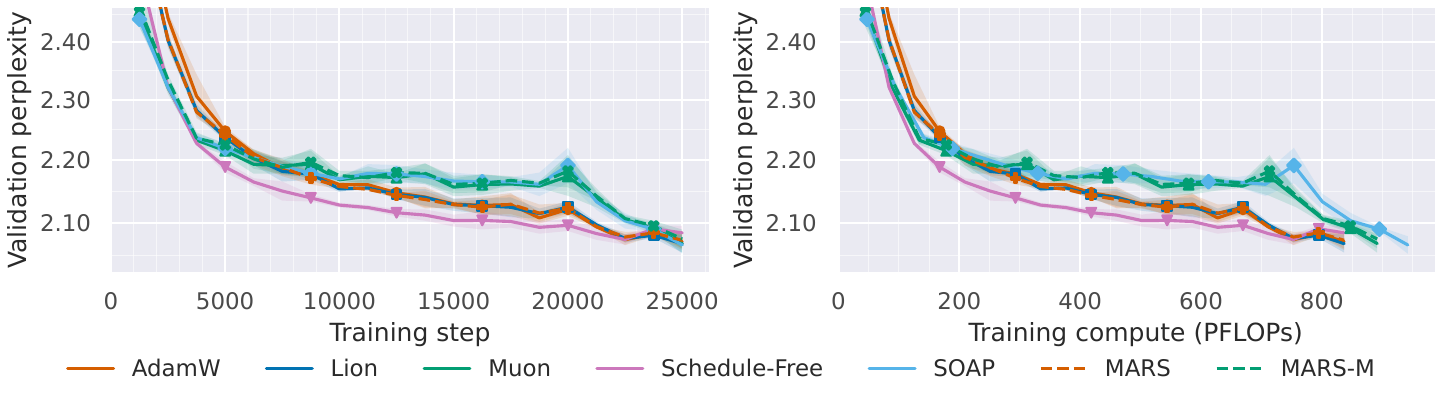}
\caption{QM9 validation perplexity on a log scale, mean over three seeds with
a $\pm1$ standard deviation band, against training steps (left) and cumulative
training compute (right). The FLOP counts differ by at most 13\% across
optimizers; per-step wall-clock does not follow FLOPs
(Table~\ref{tab:cost}, Appendix~\ref{app:cost}).}
\label{fig:qm9}
\end{figure}

QM9 and LM1B are the two tasks where AdamW is beaten by margins the paired
test resolves. In perplexity, the unit UDLM uses, SOAP finishes at 2.066,
Lion at 2.068 and Muon at 2.069 against AdamW's 2.072 (Table~\ref{tab:main}).
The gaps are 0.1--0.3\%, but they resolve because the seed moves all seven
optimizers together. MARS-M sits behind AdamW, reversing its text8 placing,
and Schedule-Free repeats its crossover, leading from 3.75k through 21.25k
steps before the decay phase passes it (Figure~\ref{fig:qm9}); it ends last
again.

Perplexity is a likelihood bound, so we also check sample quality
(Table~\ref{tab:qm9-quality}, Appendix~\ref{app:samples}). Validity,
uniqueness and mean drug-likeness are flat across optimizers, and novelty is
low for all of them because QM9 enumerates essentially every stable molecule
with up to nine heavy atoms. Frechet ChemNet Distance \citep{preuer2018fcd},
the molecular analogue of FID, is the one graded metric, and it disagrees with
the likelihood bound at $\tau = \qmNineTauFcd$: \qmNineFcdBest{} finishes
last of seven on perplexity and first on FCD. The loss-versus-samples
disagreement on images (Section~\ref{sec:celeba}) is therefore not an artifact
of image metrics.

\subsection{LM1B}
\label{sec:lm1b}

\begin{figure}[t]
\centering
\includegraphics[width=\textwidth]{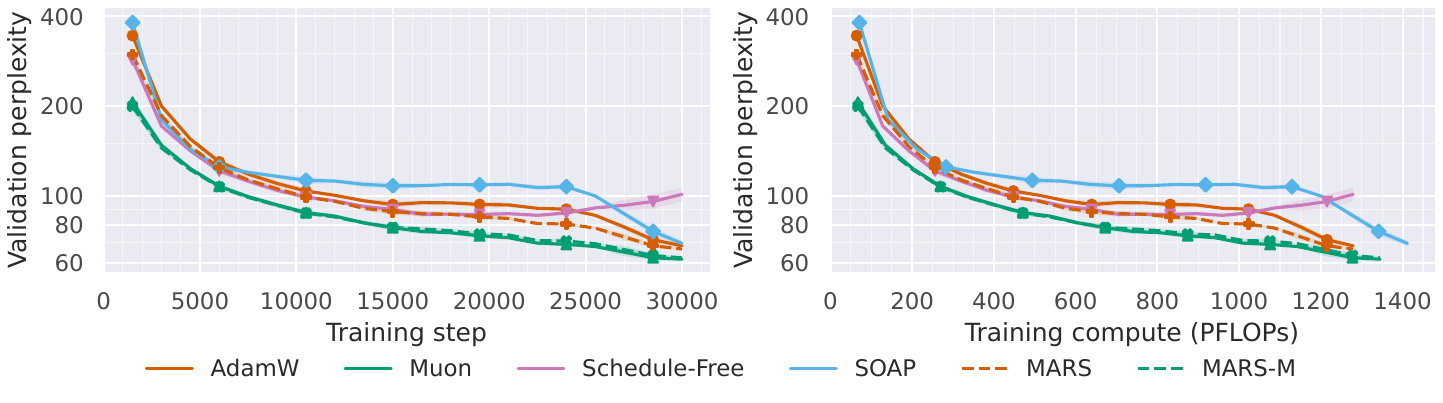}
\caption{LM1B validation perplexity on a log scale, mean over three seeds
with a $\pm1$ standard deviation band, against training steps (left) and
cumulative training compute (right). Lion is omitted: it diverges after DUO's
curriculum switch at step 15k on two of three seeds, ending near $10^4$
perplexity, and plotting it would flatten every other curve.}
\label{fig:lm1b}
\end{figure}

LM1B is the task where AdamW loses most clearly, and the task where an
optimizer breaks. Muon finishes at perplexity 61.6 and MARS-M at 62.2 against AdamW's 68.2
(Figure~\ref{fig:lm1b}),
margins of six to seven points that the paired test resolves easily; MARS also
resolves ahead (66.6), the first task where the MARS correction pays. SOAP, in first place on QM9, is behind AdamW here.

Lion fails: all three seeds track the field until step 15k, where DUO's
Gaussian curriculum hands over to the discrete objective, then two of the
three seeds climb past $10^3$ perplexity while the third recovers to 91.
Table~\ref{tab:main} marks the entry as diverged, since a mean over the seeds
($7940 \pm 12800$) would be dominated by the failed pair. Schedule-Free is
again at the back of the stable field, 33 points behind AdamW.

One caveat applies to the whole LM1B column. Our 30k steps are 3\% of the
1M-step budget of the MDLM and DUO papers (Appendix~\ref{app:budgets}), the
models are far from converged, and the leading curves are still descending at
the end of training (Figure~\ref{fig:lm1b}). Whether this ordering holds at
the full budget is left for future work.

\subsection{CelebA-64: the loss and the samples disagree}
\label{sec:celeba}

\begin{figure}[t]
\centering
\includegraphics[width=\textwidth]{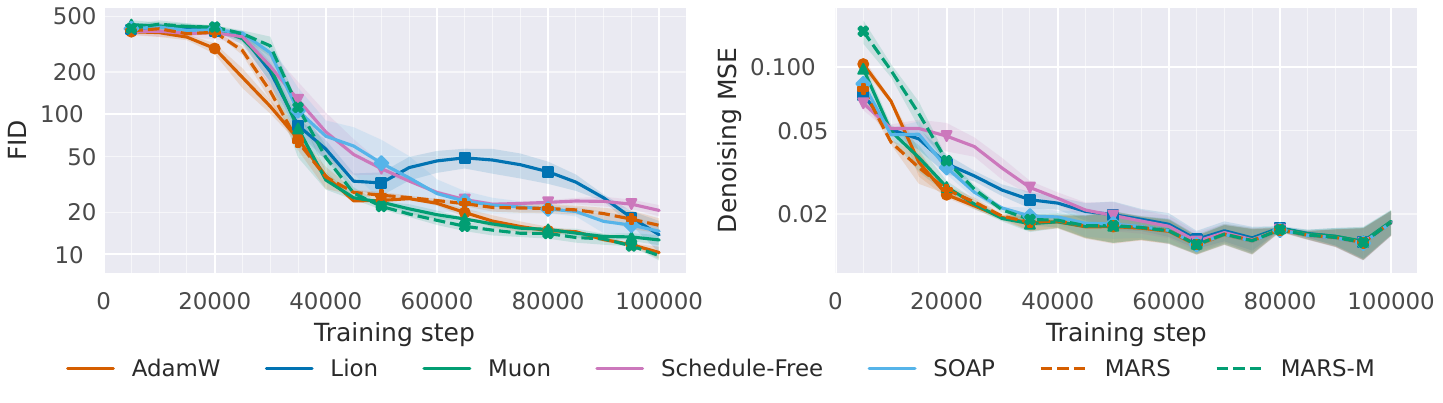}
\caption{CelebA-64 monitor FID (left) and held-out denoising MSE (right)
during training, mean over three seeds. Both are measured on the same runs:
FID separates the optimizers while the MSE curves collapse onto each other.
Lion's FID is not monotone between 50k and 65k steps.}
\label{fig:celeba}
\end{figure}

CelebA-64 is the Gaussian formulation the discrete methods descend from, and
the task with the most established sample-quality metric. The metrics disagree: the denoising loss spans just
\celebaMseSpread\% from best to worst, while the headline FID (50k samples,
DDIM-100 \citep{song2021ddim}, EMA weights; the training monitor in
Figure~\ref{fig:celeba} uses 10k samples at 50 steps, Appendix~\ref{app:eval})
runs from 6.3 to 15.3 and does not follow the loss ordering
($\tau = \celebaTau$). SOAP has the lowest loss and the
fourth-best FID; Schedule-Free is third on loss and last on FID
\citep[cf.][]{schaipp2025diffusion}, the long-standing warning of
\citet{theis2016note} that a lower likelihood-style objective need not mean
better samples. Paired against AdamW on the headline FID, five optimizers are
worse and none is better; MARS-M has the lowest mean, 6.32 against AdamW's
6.88, but the paired difference, $-0.57 \pm 0.57$, does not separate them.
The monitor and headline orderings agree except that Lion and SOAP swap fourth
and fifth, and AdamW reaches low FID earliest, 182.8 by 25k steps against
above 284 for every other optimizer.

\subsection{Cost is not what the FLOP count says}

The matrix-preconditioner methods are cheap in arithmetic and expensive in
practice (Table~\ref{tab:cost}, Appendix~\ref{app:cost}). On the CelebA UNet, SOAP adds 0.3\% to the
FLOP count of a step but 190\% to its wall-clock time, and Muon and MARS-M add
0.6\% and 104\%: their orthogonalization and eigendecomposition work happens
in many small matrix operations on convolution kernels, which use the GPU
poorly, a cost a FLOP count alone would hide
\citep{dehghani2022efficiency,kaddour2023notrain}. On the transformers the
same methods run 8\% (Muon on QM9) to 52\% (SOAP on QM9, where the $768$-wide
matrices make its eigendecompositions larger) over AdamW per step. Peak memory
is within 3\% everywhere except SOAP on QM9, where the preconditioners add
10\%.

\subsection{Rankings do not transfer across formulations}
\label{sec:transfer}

The transfer question has a negative answer. The Kendall correlation between
the final orderings of the six task pairs runs from
$\tau = \crossTauLMOneBQMNine$ (QM9 against LM1B) to
$\crossTauCelebaSixtyFourTextEight$ (text8 against CelebA-64), with five of
the six pairs at or below $+0.43$. The most discordant pair, QM9 against LM1B,
is a pair of \emph{language} tasks, so the failure to transfer is not a
text-versus-image artifact. Concretely, SOAP is second on text8, first on QM9,
fifth on LM1B and fourth on CelebA-64; MARS-M is first, sixth, second and
first; Lion is fifth, second, last and fifth: no optimizer carries its rank
across the four formulations. The consistent findings are
weaker and more useful: the tuned-AdamW baseline is beaten by a resolved
margin only on QM9 and LM1B, Schedule-Free finishes last on all four tasks,
and Lion sits near the bottom everywhere except QM9, where it is second.

\section{Discussion}
\label{sec:discussion}

\paragraph{A tuned baseline is hard to beat.} On text8 and CelebA-64 no
alternative beats AdamW once AdamW is tuned with the same care; on QM9 the
three that beat it do so by 0.003 to 0.006 perplexity, and only on
LM1B do the margins grow real (Muon and MARS-M by six to seven perplexity). This is the
pattern \citet{lucic2018gans} found for generative models and
\citet{choi2019empirical} for optimizers: much of a reported gain is the
baseline's tuning deficit. The clear negative results are Schedule-Free, last
on every task, and Lion, which diverges after LM1B's
curriculum switch on two of three seeds; the clear cost result is that the
matrix methods' overhead shows up in wall-clock, not FLOPs.

\paragraph{Scope.} We test one model size and one batch size per task, at 16.6M
to 137M parameters; optimizer gaps shrink with scale
\citep{wen2025fantastic} and depend on batch size
\citep{shallue2019dataparallelism}, so the ordering here is a statement about
this regime. Implementations follow the published algorithms, with one documented
deviation, the MARS ports omit the norm clip, and with Muon in its plain,
unchunked form (Appendix~\ref{app:optimizers}). Sample quality is graded only where a standard
metric exists, FID on CelebA-64 and FCD on QM9, and both disagree with the
loss; whether the same holds for natural-language text is untested
\citep{pillutla2021mauve}.

\paragraph{Follow-up work.} Three measurements would explain these rankings
rather than report them. The first is the smoothness along each optimizer's
trajectory, in the generalized $(L_0, L_1)$ sense of \citet{zhang2020clipping}
\citep{li2023generalized,gorbunov2024l0l1}, the regime in which gradient
clipping \citep{zhang2020clipping,koloskova2023clipping} and sign-based
updates \citep{crawshaw2022signsgd} have been analysed and from which
\citet{riabinin2026warmup} derive the warmup length; it would test whether the
tasks where AdamW is beaten have the friendlier curvature profile. The second
is a batch-size slice of the benchmark to isolate the Schedule-Free effect
\citep{zhang2025criticalbatch}, run together with the newer schedule-free
variants \citep{defazio2026schedulefreeplus}. The third is LM1B at its full
1M-step budget, without which its ordering is provisional
(Section~\ref{sec:lm1b}).

\section{Conclusion}

We benchmarked seven optimizers across four diffusion formulations under one
tuning protocol, with every winner retrained at full budget over three seeds.
A carefully tuned AdamW is a good default: no alternative beats it on text8 or
CelebA-64, and on QM9 the three that do win by under 0.3\% in perplexity. The
tuning is what makes it good, though, and the alternatives deserve the same
attention: on LM1B, Muon and MARS-M finish six to seven perplexity points
ahead of AdamW, and Muon, MARS-M and SOAP each beat it on at least one
formulation, while no optimizer wins everywhere and the ranking changes from
one formulation to the next. The lesson for practitioners is to try the
alternatives on the formulation at hand rather than inherit AdamW, and to
judge them by a sample metric as well as the loss, since the two can disagree.

\bibliographystyle{plainnat}
\bibliography{refs}

@inproceedings{loshchilov2019adamw,
  title={Decoupled Weight Decay Regularization},
  author={Ilya Loshchilov and Frank Hutter},
  booktitle={7th International Conference on Learning Representations, {ICLR} 2019, New Orleans, LA, USA, May 6-9, 2019},
  year={2019},
  url={https://openreview.net/forum?id=Bkg6RiCqY7},
}

@inproceedings{chen2023lion,
  title={Symbolic Discovery of Optimization Algorithms},
  author={Chen, Xiangning and Liang, Chen and Huang, Da and Real, Esteban and Wang, Kaiyuan and Pham, Hieu and Dong, Xuanyi and Luong, Thang and Hsieh, Cho-Jui and Lu, Yifeng and Le, Quoc V.},
  booktitle={Advances in Neural Information Processing Systems (NeurIPS)},
  year={2023}
}

@misc{jordan2024muon,
  title={Muon: An optimizer for hidden layers in neural networks},
  author={Jordan, Keller and Jin, Yuchen and Boza, Vlado and You, Jiacheng and Cesista, Franz and Newhouse, Laker and Bernstein, Jeremy},
  year={2024},
  howpublished={\url{https://kellerjordan.github.io/posts/muon/}},
}

@misc{liu2025moonlight,
  title={Muon is Scalable for LLM Training},
  author={Liu, Jingyuan and Su, Jianlin and Yao, Xingcheng and Jiang, Zhejun and Lai, Guokun and Du, Yulun and Qin, Yidao and Xu, Weixin and Lu, Enzhe and Yan, Junjie and Chen, Yanru and Zheng, Huabin and Liu, Yibo and Liu, Shaowei and Yin, Bohong and He, Weiran and Zhu, Han and Wang, Yuzhi and Wang, Jianzhou and Dong, Mengnan and Zhang, Zheng and Kang, Yongsheng and Zhang, Hao and Xu, Xinran and Zhang, Yutao and Wu, Yuxin and Zhou, Xinyu and Yang, Zhilin},
  year={2025},
  eprint={2502.16982},
  archivePrefix={arXiv},
  url={https://arxiv.org/abs/2502.16982},
}

@inproceedings{vyas2025soap,
  title={{SOAP:} Improving and Stabilizing Shampoo using Adam for Language Modeling},
  author={Nikhil Vyas and Depen Morwani and Rosie Zhao and Itai Shapira and David Brandfonbrener and Lucas Janson and Sham M. Kakade},
  booktitle={The Thirteenth International Conference on Learning Representations, {ICLR} 2025, Singapore, April 24-28, 2025},
  year={2025},
  url={https://openreview.net/forum?id=IDxZhXrpNf},
}

@inproceedings{gupta2018shampoo,
  title={Shampoo: Preconditioned Stochastic Tensor Optimization},
  author={Vineet Gupta and Tomer Koren and Yoram Singer},
  booktitle={International Conference on Machine Learning},
  pages={1842--1850},
  year={2018},
}

@inproceedings{yuan2025mars,
  title={{MARS}: Unleashing the Power of Variance Reduction for Training Large Models},
  author={Yuan, Huizhuo and Liu, Yifeng and Wu, Shuang and Zhou, Xun and Gu, Quanquan},
  booktitle={International Conference on Machine Learning (ICML)},
  year={2025}
}

@misc{liu2025marsm,
  title={{MARS-M}: When Variance Reduction Meets Matrices},
  author={Liu, Yifeng and Yuan, Angela and Gu, Quanquan},
  year={2025},
  eprint={2510.21800},
  archivePrefix={arXiv},
  url={https://arxiv.org/abs/2510.21800},
}

@inproceedings{defazio2024schedulefree,
  title={The Road Less Scheduled},
  author={Defazio, Aaron and Yang, Xingyu and Mehta, Harsh and Mishchenko, Konstantin and Khaled, Ahmed and Cutkosky, Ashok},
  booktitle={Advances in Neural Information Processing Systems 37},
  pages={9974--10007},
  publisher={Neural Information Processing Systems Foundation, Inc. (NeurIPS)},
  year={2024},
  doi={10.52202/079017-0320},
  url={https://doi.org/10.52202/079017-0320},
}

@inproceedings{ho2020ddpm,
  title={Denoising Diffusion Probabilistic Models},
  author={Ho, Jonathan and Jain, Ajay and Abbeel, Pieter},
  booktitle={Advances in Neural Information Processing Systems (NeurIPS)},
  year={2020},
}

@inproceedings{song2021ddim,
  title={Denoising Diffusion Implicit Models},
  author={Song, Jiaming and Meng, Chenlin and Ermon, Stefano},
  booktitle={International Conference on Learning Representations (ICLR)},
  year={2021},
}

@inproceedings{sahoo2024mdlm,
  title={Simple and Effective Masked Diffusion Language Models},
  author={Sahoo, Subham and Arriola, Marianne and Schiff, Yair and Gokaslan, Aaron and Marroquin, Edgar and Chiu, Justin and Rush, Alexander and Kuleshov, Volodymyr},
  booktitle={Advances in Neural Information Processing Systems 37},
  pages={130136--130184},
  publisher={Neural Information Processing Systems Foundation, Inc. (NeurIPS)},
  year={2024},
  doi={10.52202/079017-4135},
  url={https://doi.org/10.52202/079017-4135},
}

@inproceedings{schiff2025guidance,
  title={Simple Guidance Mechanisms for Discrete Diffusion Models},
  author={Schiff, Yair and Sahoo, Subham Sekhar and Phung, Hao and Wang, Guanghan and Boshar, Sam and Dalla-torre, Hugo and de Almeida, Bernardo P. and Rush, Alexander and Pierrot, Thomas and Kuleshov, Volodymyr},
  booktitle={International Conference on Learning Representations (ICLR)},
  year={2025},
}

@inproceedings{sahoo2025duo,
  title={The Diffusion Duality},
  author={Sahoo, Subham Sekhar and Deschenaux, Justin and Gokaslan, Aaron and Wang, Guanghan and Chiu, Justin and Kuleshov, Volodymyr},
  booktitle={International Conference on Machine Learning (ICML)},
  year={2025},
}

@inproceedings{lou2024sedd,
  title={Discrete Diffusion Modeling by Estimating the Ratios of the Data Distribution},
  author={Lou, Aaron and Meng, Chenlin and Ermon, Stefano},
  booktitle={International Conference on Machine Learning (ICML)},
  year={2024},
}

@inproceedings{austin2021d3pm,
  title={Structured Denoising Diffusion Models in Discrete State-Spaces},
  author={Austin, Jacob and Johnson, Daniel D. and Ho, Jonathan and Tarlow, Daniel and van den Berg, Rianne},
  booktitle={Advances in Neural Information Processing Systems (NeurIPS)},
  year={2021},
}

@inproceedings{gong2025diffullama,
  title={Scaling Diffusion Language Models via Adaptation from Autoregressive Models},
  author={Gong, Shansan and Agarwal, Shivam and Zhang, Yizhe and Ye, Jiacheng and Zheng, Lin and Li, Mukai and An, Chenxin and Zhao, Peilin and Bi, Wei and Han, Jiawei and Peng, Hao and Kong, Lingpeng},
  booktitle={International Conference on Learning Representations (ICLR)},
  year={2025},
}

@misc{vonrutte2025scaling,
  title={Scaling Behavior of Discrete Diffusion Language Models},
  author={von R{\"u}tte, Dimitri and Fluri, Janis and Pooladzandi, Omead and Sch{\"o}lkopf, Bernhard and Hofmann, Thomas and Orvieto, Antonio},
  year={2025},
  eprint={2512.10858},
  archivePrefix={arXiv},
  url={https://arxiv.org/abs/2512.10858},
}

@misc{semenov2025benchmarking,
  title={Benchmarking Optimizers for Large Language Model Pretraining},
  author={Semenov, Andrei and Pagliardini, Matteo and Jaggi, Martin},
  year={2025},
  eprint={2509.01440},
  archivePrefix={arXiv},
  url={https://arxiv.org/abs/2509.01440},
}

@misc{wen2025fantastic,
  title={Fantastic Pretraining Optimizers and Where to Find Them},
  author={Wen, Kaiyue and Hall, David and Ma, Tengyu and Liang, Percy},
  year={2025},
  eprint={2509.02046},
  archivePrefix={arXiv},
  url={https://arxiv.org/abs/2509.02046},
}

@misc{schaipp2025diffusion,
  title={Optimization Benchmark for Diffusion Models on Dynamical Systems},
  author={Schaipp, Fabian},
  year={2025},
  eprint={2510.19376},
  archivePrefix={arXiv},
  url={https://arxiv.org/abs/2510.19376},
}

@misc{chen2026cmuon,
  title={CMuon: Accelerating and Stabilizing Diffusion Transformer Training via Chunked Momentum Orthogonalization},
  author={Chen, Chuyan and Sun, Peng and Yuan, Kun},
  year={2026},
  eprint={2608.02502},
  archivePrefix={arXiv},
  url={https://arxiv.org/abs/2608.02502},
}

@misc{dahl2023algoperf,
  title={Benchmarking Neural Network Training Algorithms},
  author={Dahl, George E. and Schneider, Frank and Nado, Zachary and Agarwal, Naman and Sastry, Chandramouli Shama and Hennig, Philipp and Medapati, Sourabh and Eschenhagen, Runa and Kasimbeg, Priya and Suo, Daniel and Bae, Juhan and Gilmer, Justin and Peirson, Abel L. and Khan, Bilal and Anil, Rohan and Rabbat, Mike and Krishnan, Shankar and Snider, Daniel and Amid, Ehsan and Chen, Kongtao and Maddison, Chris J. and Vasudev, Rakshith and Badura, Michal and Garg, Ankush and Mattson, Peter},
  year={2023},
  eprint={2306.07179},
  archivePrefix={arXiv},
  url={https://arxiv.org/abs/2306.07179},
}

@inproceedings{akiba2019optuna,
  title={Optuna: A Next-generation Hyperparameter Optimization Framework},
  author={Akiba, Takuya and Sano, Shotaro and Yanase, Toshihiko and Ohta, Takeru and Koyama, Masanori},
  booktitle={ACM SIGKDD International Conference on Knowledge Discovery and Data Mining (KDD)},
  year={2019},
}

@article{li2018hyperband,
  title={Hyperband: A Novel Bandit-Based Approach to Hyperparameter Optimization},
  author={Li, Lisha and Jamieson, Kevin and DeSalvo, Giulia and Rostamizadeh, Afshin and Talwalkar, Ameet},
  journal={Journal of Machine Learning Research},
  volume={18},
  pages={1--52},
  year={2018}
}

@inproceedings{bergstra2011tpe,
  title={Algorithms for Hyper-Parameter Optimization},
  author={Bergstra, James and Bardenet, R{\'e}mi and Bengio, Yoshua and K{\'e}gl, Bal{\'a}zs},
  booktitle={Advances in Neural Information Processing Systems (NIPS)},
  year={2011},
}

@misc{mahoney2011text8,
  title={Large Text Compression Benchmark},
  author={Mahoney, Matt},
  year={2011},
  howpublished={\url{http://mattmahoney.net/dc/text.html}},
}

@inproceedings{chelba2013lm1b,
  title={One billion word benchmark for measuring progress in statistical language modeling},
  author={Chelba, Ciprian and Mikolov, Tomas and Schuster, Mike and Ge, Qi and Brants, Thorsten and Koehn, Phillipp and Robinson, Tony},
  booktitle={Interspeech 2014},
  pages={2635--2639},
  publisher={ISCA},
  year={2014},
  doi={10.21437/interspeech.2014-564},
  url={https://doi.org/10.21437/interspeech.2014-564},
}

@article{ramakrishnan2014qm9,
  title={Quantum chemistry structures and properties of 134 kilo molecules},
  author={Ramakrishnan, Raghunathan and Dral, Pavlo O. and Rupp, Matthias and von Lilienfeld, O. Anatole},
  journal={Scientific Data},
  volume={1},
  number={1},
  publisher={Springer Science and Business Media LLC},
  year={2014},
  doi={10.1038/sdata.2014.22},
  url={https://doi.org/10.1038/sdata.2014.22},
}

@inproceedings{liu2015celeba,
  title={Deep Learning Face Attributes in the Wild},
  author={Liu, Ziwei and Luo, Ping and Wang, Xiaogang and Tang, Xiaoou},
  booktitle={2015 {IEEE} International Conference on Computer Vision ({ICCV})},
  pages={3730--3738},
  publisher={IEEE},
  year={2015},
  doi={10.1109/iccv.2015.425},
  url={https://doi.org/10.1109/iccv.2015.425},
}

@inproceedings{hu2024minicpm,
  title={{MiniCPM}: Unveiling the Potential of Small Language Models with Scalable Training Strategies},
  author={Hu, Shengding and Tu, Yuge and Han, Xu and He, Chaoqun and Cui, Ganqu and Long, Xiang and Zheng, Zhi and Fang, Yewei and Huang, Yuxiang and Zhao, Weilin and Zhang, Xinrong and Thai, Zheng Leng and Zhang, Kaihuo and Wang, Chongyi and Yao, Yuan and Zhao, Chenyang and Zhou, Jie and Cai, Jie and Zhai, Zhongwu and Ding, Ning and Jia, Chao and Zeng, Guoyang and Li, Dahai and Liu, Zhiyuan and Sun, Maosong},
  booktitle={Conference on Language Modeling (COLM)},
  year={2024},
}

@inproceedings{hagele2024scaling,
  title={Scaling Laws and Compute-Optimal Training Beyond Fixed Training Durations},
  author={H{\"a}gele, Alexander and Bakouch, Elie and Kosson, Atli and Ben Allal, Loubna and Von Werra, Leandro and Jaggi, Martin},
  booktitle={Advances in Neural Information Processing Systems (NeurIPS)},
  year={2024}
}

@inproceedings{shi2024md4,
  title={Simplified and Generalized Masked Diffusion for Discrete Data},
  author={Shi, Jiaxin and Han, Kehang and Wang, Zhe and Doucet, Arnaud and Titsias, Michalis},
  booktitle={Advances in Neural Information Processing Systems 37},
  pages={103131--103167},
  publisher={Neural Information Processing Systems Foundation, Inc. (NeurIPS)},
  year={2024},
  doi={10.52202/079017-3277},
  url={https://doi.org/10.52202/079017-3277},
}

@misc{riabinin2026warmup,
  title={Where Does Warm-Up Come From? Adaptive Scheduling for Norm-Constrained Optimizers},
  author={Riabinin, Artem and Veprikov, Andrey and Bolatov, Arman and Tak\'{a}\v{c}, Martin and Beznosikov, Aleksandr},
  year={2026},
  eprint={2602.05813},
  archivePrefix={arXiv},
  url={https://arxiv.org/abs/2602.05813},
}

@misc{veprikov2025preconditioned,
  title={Preconditioned Norms: A Unified Framework for Steepest Descent, Quasi-Newton and Adaptive Methods},
  author={Veprikov, Andrey and Bolatov, Arman and Bogdanov, Aleksandr and Horv\'{a}th, Samuel and Beznosikov, Aleksandr and Tak\'{a}\v{c}, Martin and Hanzely, Slavomir},
  year={2025},
  eprint={2510.10777},
  archivePrefix={arXiv},
  url={https://arxiv.org/abs/2510.10777},
}

@inproceedings{cutkosky2019storm,
  title     = {Momentum-Based Variance Reduction in Non-Convex {SGD}},
  author    = {Cutkosky, Ashok and Orabona, Francesco},
  booktitle = {Advances in Neural Information Processing Systems},
  year      = {2019},
}

@misc{bernstein2024norm,
  title={Old Optimizer, New Norm: An Anthology},
  author={Bernstein, Jeremy and Newhouse, Laker},
  year={2024},
  eprint={2409.20325},
  archivePrefix={arXiv},
  url={https://arxiv.org/abs/2409.20325},
}

@article{bjorck1971orthogonal,
  title={An Iterative Algorithm for Computing the Best Estimate of an Orthogonal Matrix},
  author={Bj\"orck, {\AA}. and Bowie, C.},
  journal={{SIAM} Journal on Numerical Analysis},
  volume={8},
  number={2},
  pages={358--364},
  publisher={Society for Industrial \& Applied Mathematics (SIAM)},
  year={1971},
  doi={10.1137/0708036},
  url={https://doi.org/10.1137/0708036},
}

@inproceedings{bouthillier2021variance,
 author = {Bouthillier, Xavier and Delaunay, Pierre and Bronzi, Mirko and Trofimov, Assya and Nichyporuk, Brennan and Szeto, Justin and Mohammadi Sepahvand, Nazanin and Raff, Edward and Madan, Kanika and Voleti, Vikram and Ebrahimi Kahou, Samira and Michalski, Vincent and Arbel, Tal and Pal, Chris and Varoquaux, Gael and Vincent, Pascal},
 booktitle = {Proceedings of Machine Learning and Systems},
 editor = {A. Smola and A. Dimakis and I. Stoica},
 pages = {747--769},
 title = {Accounting for Variance in Machine Learning Benchmarks},
 url = {https://proceedings.mlsys.org/paper_files/paper/2021/file/0184b0cd3cfb185989f858a1d9f5c1eb-Paper.pdf},
 volume = {3},
 year = {2021}
}

@inproceedings{campbell2022continuous,
  title        = {A Continuous Time Framework for Discrete Denoising Models},
  author       = {Campbell, Andrew and Benton, Joe and De Bortoli, Valentin and Rainforth, Thomas and Deligiannidis, George and Doucet, Arnaud},
  booktitle    = {Advances in Neural Information Processing Systems},
  volume       = {35},
  pages        = {28266--28279},
  publisher    = {Curran Associates, Inc.},
  year         = {2022},
  eprint       = {2205.14987},
  archivePrefix= {arXiv},
  primaryClass = {stat.ML}
}

@misc{choi2019empirical,
  title={On Empirical Comparisons of Optimizers for Deep Learning},
  author={Choi, Dami and Shallue, Christopher J. and Nado, Zachary and Lee, Jaehoon and Maddison, Chris J. and Dahl, George E.},
  year={2019},
  eprint={1910.05446},
  archivePrefix={arXiv},
  url={https://arxiv.org/abs/1910.05446},
}

@inproceedings{dehghani2022efficiency,
  title={The Efficiency Misnomer},
  author={Mostafa Dehghani and Yi Tay and Anurag Arnab and Lucas Beyer and Ashish Vaswani},
  booktitle={10th International Conference on Learning Representations ({ICLR})},
  publisher={OpenReview.net},
  year={2022},
  eprint={2110.12894},
  archivePrefix={arXiv},
}

@inproceedings{devlin2019bert,
  title={{BERT:} Pre-training of Deep Bidirectional Transformers for Language Understanding},
  author={Jacob Devlin and Ming{-}Wei Chang and Kenton Lee and Kristina Toutanova},
  editor={Jill Burstein and Christy Doran and Thamar Solorio},
  booktitle={Proceedings of the 2019 Conference of the North American Chapter of the Association for Computational Linguistics: Human Language Technologies, {NAACL-HLT} 2019, Minneapolis, MN, USA, June 2-7, 2019, Volume 1 (Long and Short Papers)},
  pages={4171--4186},
  publisher={Association for Computational Linguistics},
  year={2019},
  doi={10.18653/v1/n19-1423},
  url={https://doi.org/10.18653/v1/n19-1423},
}

@inproceedings{dodge2019showyourwork,
  title={Show Your Work: Improved Reporting of Experimental Results},
  author={Dodge, Jesse and Gururangan, Suchin and Card, Dallas and Schwartz, Roy and Smith, Noah A.},
  booktitle={Proceedings of the 2019 Conference on Empirical Methods in Natural Language Processing and the 9th International Joint Conference on Natural Language Processing ({EMNLP}-{IJCNLP})},
  pages={2185--2194},
  publisher={Association for Computational Linguistics},
  year={2019},
  doi={10.18653/v1/d19-1224},
  url={https://doi.org/10.18653/v1/d19-1224},
}

@article{duchi2011adagrad,
  title   = {Adaptive Subgradient Methods for Online Learning and Stochastic Optimization},
  author  = {Duchi, John and Hazan, Elad and Singer, Yoram},
  journal = {Journal of Machine Learning Research},
  volume  = {12},
  pages   = {2121--2159},
  year    = {2011}
}

@inproceedings{heusel2017fid,
  title         = {{GANs} Trained by a Two Time-Scale Update Rule Converge to a Local Nash Equilibrium},
  author        = {Heusel, Martin and Ramsauer, Hubert and Unterthiner, Thomas and Nessler, Bernhard and Hochreiter, Sepp},
  booktitle     = {Advances in Neural Information Processing Systems},
  year          = {2017},
  eprint        = {1706.08500},
  archivePrefix = {arXiv}
}

@book{higham2008functions,
  title={Functions of Matrices: Theory and Computation},
  author={Higham, Nicholas J.},
  publisher={Society for Industrial and Applied Mathematics},
  address={Philadelphia, PA, USA},
  year={2008},
  doi={10.1137/1.9780898717778},
  isbn={978-0-898716-46-7},
}

@article{holm1979sequential,
  title={A Simple Sequentially Rejective Multiple Test Procedure},
  author={Sture Holm},
  journal={Scandinavian Journal of Statistics},
  volume={6},
  number={2},
  pages={65--70},
  year={1979},
  url={https://www.jstor.org/stable/4615733},
}

@inproceedings{hoogeboom2021argmax,
  title        = {Argmax Flows and Multinomial Diffusion: Learning Categorical Distributions},
  author       = {Hoogeboom, Emiel and Nielsen, Didrik and Jaini, Priyank and Forr\'{e}, Patrick and Welling, Max},
  booktitle    = {Advances in Neural Information Processing Systems},
  volume       = {34},
  pages        = {12454--12465},
  publisher    = {Curran Associates, Inc.},
  year         = {2021},
  eprint       = {2102.05379},
  archivePrefix= {arXiv},
  primaryClass = {stat.ML}
}

@article{jelinek1977perplexity,
  title={Perplexity---a measure of the difficulty of speech recognition tasks},
  author={Jelinek, F. and Mercer, R. L. and Bahl, L. R. and Baker, J. K.},
  journal={The Journal of the Acoustical Society of America},
  volume={62},
  number={S1},
  pages={S63--S63},
  publisher={Acoustical Society of America (ASA)},
  year={1977},
  doi={10.1121/1.2016299},
  url={https://doi.org/10.1121/1.2016299},
}

@inproceedings{kaddour2023notrain,
  title={No Train No Gain: Revisiting Efficient Training Algorithms For Transformer-based Language Models},
  author={Kaddour, Jean and Key, Oscar and Nawrot, Piotr and Minervini, Pasquale and Kusner, Matt},
  booktitle={Advances in Neural Information Processing Systems 36},
  pages={25793--25818},
  publisher={Neural Information Processing Systems Foundation, Inc. (NeurIPS)},
  year={2023},
  doi={10.52202/075280-1122},
  url={https://doi.org/10.52202/075280-1122},
}

@inproceedings{kasimbeg2025algoperf,
  title        = {Accelerating Neural Network Training: An Analysis of the {AlgoPerf} Competition},
  author       = {Kasimbeg, Priya and Schneider, Frank and Eschenhagen, Runa and Bae, Juhan and {Shama Sastry}, Chandramouli and Saroufim, Mark and Feng, Boyuan and Wright, Less and Yang, Edward Z. and Nado, Zachary and Medapati, Sourabh and Hennig, Philipp and Rabbat, Michael and Dahl, George E.},
  booktitle    = {International Conference on Learning Representations (ICLR)},
  year         = {2025},
  eprint       = {2502.15015},
  archivePrefix= {arXiv},
  primaryClass = {cs.LG}
}

@misc{kim2021pytorchoptimizer,
  title        = {pytorch\_optimizer: optimizer \& lr scheduler \& loss function collections in {PyTorch}},
  author       = {Kim, Hyeongchan},
  year         = {2021},
  url          = {https://github.com/kozistr/pytorch_optimizer}
}

@inproceedings{kingma2015adam,
  title        = {Adam: A Method for Stochastic Optimization},
  author       = {Kingma, Diederik P. and Ba, Jimmy},
  booktitle    = {International Conference on Learning Representations (ICLR)},
  year         = {2015},
  eprint       = {1412.6980},
  archivePrefix= {arXiv},
  primaryClass = {cs.LG}
}

@inproceedings{kingma2021vdm,
  title     = {Variational Diffusion Models},
  author    = {Kingma, Diederik P. and Salimans, Tim and Poole, Ben and Ho, Jonathan},
  booktitle = {Advances in Neural Information Processing Systems},
  volume    = {34},
  pages     = {21696--21707},
  year      = {2021},
  eprint    = {2107.00630},
  archivePrefix = {arXiv}
}

@inproceedings{krogh1991weight,
  title={A Simple Weight Decay Can Improve Generalization},
  author={Krogh, Anders and Hertz, John A.},
  booktitle={Advances in Neural Information Processing Systems 4 (NIPS)},
  pages={950--957},
  publisher={Morgan Kaufmann},
  year={1991},
}

@inproceedings{lucic2018gans,
  title={Are {GANs} Created Equal? A Large-Scale Study},
  author={Lucic, Mario and Kurach, Karol and Michalski, Marcin and Gelly, Sylvain and Bousquet, Olivier},
  booktitle={Advances in Neural Information Processing Systems (NeurIPS)},
  pages={698--707},
  year={2018},
}

@inproceedings{micikevicius2018mixed,
  title={Mixed Precision Training},
  author={Micikevicius, Paulius and Narang, Sharan and Alben, Jonah and Diamos, Gregory and Elsen, Erich and Garcia, David and Ginsburg, Boris and Houston, Michael and Kuchaiev, Oleksii and Venkatesh, Ganesh and Wu, Hao},
  booktitle={International Conference on Learning Representations (ICLR)},
  year={2018},
  eprint={1710.03740},
  archivePrefix={arXiv},
}

@techreport{mikolov2012subword,
  title={Subword language modeling with neural networks},
  author={Mikolov, Tom\'{a}\v{s} and Sutskever, Ilya and Deoras, Anoop and Le, Hai-Son and Kombrink, Stefan and \v{C}ernock\'{y}, Jan},
  year={2012},
  url={http://www.fit.vutbr.cz/~imikolov/rnnlm/char.pdf},
  institution={Brno University of Technology},
  note={Unpublished manuscript},
}

@misc{obukhov2020torchfidelity,
  title={High-fidelity performance metrics for generative models in {PyTorch}},
  author={Obukhov, Anton and Seitzer, Maximilian and Wu, Po-Wei and Zhydenko, Semen and Kyl, Jonathan and Lin, Elvis Yu-Jing},
  publisher={Zenodo},
  year={2020},
  doi={10.5281/zenodo.3786539},
  howpublished={\url{https://github.com/toshas/torch-fidelity}},
  version={v0.4.0},
}

@inproceedings{pagliardini2025ademamix,
  title={The {AdEMAMix} Optimizer: Better, Faster, Older},
  author={Pagliardini, Matteo and Ablin, Pierre and Grangier, David},
  booktitle={International Conference on Learning Representations (ICLR)},
  year={2025},
  eprint={2409.03137},
  archivePrefix={arXiv},
  primaryClass={cs.LG},
  url={https://openreview.net/forum?id=jj7b3p5kLY},
}

@inproceedings{parmar2022cleanfid,
  title={On Aliased Resizing and Surprising Subtleties in {GAN} Evaluation},
  author={Parmar, Gaurav and Zhang, Richard and Zhu, Jun-Yan},
  booktitle={2022 {IEEE/CVF} Conference on Computer Vision and Pattern Recognition ({CVPR})},
  pages={11400--11410},
  publisher={IEEE},
  year={2022},
  doi={10.1109/cvpr52688.2022.01112},
  url={https://doi.org/10.1109/cvpr52688.2022.01112},
}

@inproceedings{pascanu2013difficulty,
  title={On the difficulty of training recurrent neural networks},
  author={Pascanu, Razvan and Mikolov, Tomas and Bengio, Yoshua},
  booktitle={Proceedings of the 30th International Conference on Machine Learning},
  series={Proceedings of Machine Learning Research},
  volume={28},
  pages={1310--1318},
  publisher={PMLR},
  year={2013},
  eprint={1211.5063},
  archivePrefix={arXiv},
}

@inproceedings{paszke2019pytorch,
  title={{PyTorch}: An Imperative Style, High-Performance Deep Learning Library},
  author={Paszke, Adam and Gross, Sam and Massa, Francisco and Lerer, Adam and Bradbury, James and Chanan, Gregory and Killeen, Trevor and Lin, Zeming and Gimelshein, Natalia and Antiga, Luca and Desmaison, Alban and K\"{o}pf, Andreas and Yang, Edward and DeVito, Zach and Raison, Martin and Tejani, Alykhan and Chilamkurthy, Sasank and Steiner, Benoit and Fang, Lu and Bai, Junjie and Chintala, Soumith},
  booktitle={Advances in Neural Information Processing Systems},
  volume={32},
  pages={8024--8035},
  year={2019},
  eprint={1912.01703},
  archivePrefix={arXiv},
}

@inproceedings{peebles2023dit,
  title={Scalable Diffusion Models with Transformers},
  author={Peebles, William and Xie, Saining},
  booktitle={2023 {IEEE/CVF} International Conference on Computer Vision ({ICCV})},
  pages={4172--4182},
  publisher={IEEE},
  year={2023},
  doi={10.1109/iccv51070.2023.00387},
  url={https://doi.org/10.1109/iccv51070.2023.00387},
}

@inproceedings{pethick2025scion,
  title={Training Deep Learning Models with Norm-Constrained {LMO}s},
  author={Pethick, Thomas and Xie, Wanyun and Antonakopoulos, Kimon and Zhu, Zhenyu and Silveti-Falls, Antonio and Cevher, Volkan},
  booktitle={Proceedings of the 42nd International Conference on Machine Learning (ICML)},
  series={Proceedings of Machine Learning Research},
  volume={267},
  pages={49069--49104},
  publisher={PMLR},
  year={2025},
  eprint={2502.07529},
  archivePrefix={arXiv},
  primaryClass={cs.LG},
}

@inproceedings{pillutla2021mauve,
  title={{MAUVE}: Measuring the Gap Between Neural Text and Human Text using Divergence Frontiers},
  author={Pillutla, Krishna and Swayamdipta, Swabha and Zellers, Rowan and Thickstun, John and Welleck, Sean and Choi, Yejin and Harchaoui, Zaid},
  editor={M. Ranzato and A. Beygelzimer and Y. Dauphin and P. S. Liang and J. Wortman Vaughan},
  booktitle={Advances in Neural Information Processing Systems},
  volume={34},
  pages={4816--4828},
  publisher={Curran Associates, Inc.},
  year={2021},
  eprint={2102.01454},
  archivePrefix={arXiv},
  primaryClass={cs.CL},
}

@article{polyak1992averaging,
  title={Acceleration of Stochastic Approximation by Averaging},
  author={Polyak, B. T. and Juditsky, A. B.},
  journal={{SIAM} Journal on Control and Optimization},
  volume={30},
  number={4},
  pages={838--855},
  publisher={Society for Industrial \& Applied Mathematics (SIAM)},
  year={1992},
  doi={10.1137/0330046},
  url={https://doi.org/10.1137/0330046},
}

@inproceedings{ronneberger2015unet,
  title         = {{U-Net}: Convolutional Networks for Biomedical Image Segmentation},
  author        = {Ronneberger, Olaf and Fischer, Philipp and Brox, Thomas},
  booktitle     = {Medical Image Computing and Computer-Assisted Intervention -- MICCAI 2015},
  series        = {Lecture Notes in Computer Science},
  volume        = {9351},
  pages         = {234--241},
  publisher     = {Springer},
  year          = {2015},
  doi           = {10.1007/978-3-319-24574-4_28},
  eprint        = {1505.04597},
  archivePrefix = {arXiv}
}

@inproceedings{schmidt2021crowded,
  title={Descending through a Crowded Valley --- Benchmarking Deep Learning Optimizers},
  author={Schmidt, Robin M. and Schneider, Frank and Hennig, Philipp},
  booktitle={Proceedings of the 38th International Conference on Machine Learning (ICML)},
  series={Proceedings of Machine Learning Research},
  volume={139},
  pages={9367--9376},
  year={2021},
  eprint={2007.01547},
  archivePrefix={arXiv},
}

@article{shallue2019dataparallelism,
  title={Measuring the Effects of Data Parallelism on Neural Network Training},
  author={Shallue, Christopher J. and Lee, Jaehoon and Antognini, Joseph and Sohl-Dickstein, Jascha and Frostig, Roy and Dahl, George E.},
  journal={Journal of Machine Learning Research},
  volume={20},
  number={112},
  pages={1--49},
  year={2019},
}

@inproceedings{sivaprasad2020tuning,
  title     = {Optimizer Benchmarking Needs to Account for Hyperparameter Tuning},
  author    = {Sivaprasad, {Prabhu Teja} and Mai, Florian and Vogels, Thijs and Jaggi, Martin and Fleuret, Fran\c{c}ois},
  booktitle = {Proceedings of the 37th International Conference on Machine Learning (ICML)},
  series    = {Proceedings of Machine Learning Research},
  volume    = {119},
  pages     = {9036--9045},
  year      = {2020}
}

@inproceedings{sohldickstein2015diffusion,
  title={Deep Unsupervised Learning using Nonequilibrium Thermodynamics},
  author={Sohl-Dickstein, Jascha and Weiss, Eric A. and Maheswaranathan, Niru and Ganguli, Surya},
  booktitle={Proceedings of the 32nd International Conference on Machine Learning (ICML)},
  series={Proceedings of Machine Learning Research},
  volume={37},
  pages={2256--2265},
  year={2015},
  eprint={1503.03585},
  archivePrefix={arXiv},
}

@inproceedings{song2019ncsn,
  title={Generative Modeling by Estimating Gradients of the Data Distribution},
  author={Song, Yang and Ermon, Stefano},
  booktitle={Advances in Neural Information Processing Systems 32 (NeurIPS)},
  year={2019},
  eprint={1907.05600},
  archivePrefix={arXiv},
  primaryClass={cs.LG},
}

@inproceedings{song2020improved,
  title={Improved Techniques for Training Score-Based Generative Models},
  author={Song, Yang and Ermon, Stefano},
  booktitle={Advances in Neural Information Processing Systems 33 (NeurIPS)},
  year={2020},
  eprint={2006.09011},
  archivePrefix={arXiv},
  primaryClass={cs.LG},
}

@inproceedings{song2021sde,
  title={Score-Based Generative Modeling through Stochastic Differential Equations},
  author={Song, Yang and Sohl-Dickstein, Jascha and Kingma, Diederik P. and Kumar, Abhishek and Ermon, Stefano and Poole, Ben},
  booktitle={International Conference on Learning Representations (ICLR)},
  year={2021}
}

@article{student1908probable,
  title   = {The Probable Error of a Mean},
  author  = {{Student}},
  journal = {Biometrika},
  volume  = {6},
  number  = {1},
  pages   = {1--25},
  year    = {1908},
  doi     = {10.1093/biomet/6.1.1},
  note    = {Published under the pseudonym ``Student'' by William Sealy Gosset}
}

@inproceedings{sutskever2013momentum,
  title={On the Importance of Initialization and Momentum in Deep Learning},
  author={Sutskever, Ilya and Martens, James and Dahl, George and Hinton, Geoffrey},
  booktitle={Proceedings of the 30th International Conference on Machine Learning (ICML)},
  series={Proceedings of Machine Learning Research},
  volume={28},
  pages={1139--1147},
  publisher={PMLR},
  year={2013},
}

@inproceedings{szegedy2016inception,
  title={Rethinking the Inception Architecture for Computer Vision},
  author={Szegedy, Christian and Vanhoucke, Vincent and Ioffe, Sergey and Shlens, Jon and Wojna, Zbigniew},
  booktitle={2016 {IEEE} Conference on Computer Vision and Pattern Recognition ({CVPR})},
  pages={2818--2826},
  publisher={IEEE},
  year={2016},
  doi={10.1109/cvpr.2016.308},
  url={https://doi.org/10.1109/cvpr.2016.308},
}

@inproceedings{theis2016note,
  title={A Note on the Evaluation of Generative Models},
  author={Theis, Lucas and van den Oord, A\"{a}ron and Bethge, Matthias},
  booktitle={International Conference on Learning Representations (ICLR)},
  year={2016},
  eprint={1511.01844},
  archivePrefix={arXiv},
}

@misc{tieleman2012rmsprop,
  title={Lecture 6.5---RMSProp: Divide the Gradient by a Running Average of its Recent Magnitude},
  author={Tieleman, Tijmen and Hinton, Geoffrey},
  year={2012},
  note={COURSERA: Neural Networks for Machine Learning},
}

@inproceedings{vaswani2017attention,
  title={Attention is All you Need},
  author={Vaswani, Ashish and Shazeer, Noam and Parmar, Niki and Uszkoreit, Jakob and Jones, Llion and Gomez, Aidan N. and Kaiser, {\L}ukasz and Polosukhin, Illia},
  booktitle={Advances in Neural Information Processing Systems},
  volume={30},
  publisher={Curran Associates, Inc.},
  year={2017},
  eprint={1706.03762},
  archivePrefix={arXiv},
  primaryClass={cs.CL},
}

@misc{vonplaten2022diffusers,
  title={Diffusers: State-of-the-art diffusion models},
  author={von Platen, Patrick and Patil, Suraj and Lozhkov, Anton and Cuenca, Pedro and Lambert, Nathan and Rasul, Kashif and Davaadorj, Mishig and Nair, Dhruv and Paul, Sayak and Berman, William and Xu, Yiyi and Liu, Steven and Wolf, Thomas},
  journal={GitHub repository},
  publisher={GitHub},
  year={2022},
  howpublished={\url{https://github.com/huggingface/diffusers}},
}

@misc{wang2023lionpytorch,
  title={lion-pytorch: {L}ion Optimizer for {PyTorch}},
  author={Wang, Phil},
  year={2023},
  howpublished={\url{https://github.com/lucidrains/lion-pytorch}},
  note={Software, version 0.2.4},
}

@misc{watanabe2023tpe,
  title={Tree-Structured Parzen Estimator: Understanding Its Algorithm Components and Their Roles for Better Empirical Performance},
  author={Watanabe, Shuhei},
  year={2023},
  eprint={2304.11127},
  archivePrefix={arXiv},
  url={https://arxiv.org/abs/2304.11127},
}

@article{weininger1988smiles,
  title={{SMILES}, a chemical language and information system. 1. Introduction to methodology and encoding rules},
  author={Weininger, David},
  journal={Journal of Chemical Information and Computer Sciences},
  volume={28},
  number={1},
  pages={31--36},
  publisher={American Chemical Society (ACS)},
  year={1988},
  doi={10.1021/ci00057a005},
  url={https://doi.org/10.1021/ci00057a005},
}

@inproceedings{wolf2020transformers,
  title     = {Transformers: State-of-the-Art Natural Language Processing},
  author    = {Wolf, Thomas and Debut, Lysandre and Sanh, Victor and Chaumond, Julien and Delangue, Clement and Moi, Anthony and Cistac, Pierric and Rault, Tim and Louf, R\'{e}mi and Funtowicz, Morgan and Davison, Joe and Shleifer, Sam and von Platen, Patrick and Ma, Clara and Jernite, Yacine and Plu, Julien and Xu, Canwen and Le Scao, Teven and Gugger, Sylvain and Drame, Mariama and Lhoest, Quentin and Rush, Alexander M.},
  booktitle = {Proceedings of the 2020 Conference on Empirical Methods in Natural Language Processing: System Demonstrations},
  pages     = {38--45},
  address   = {Online},
  publisher = {Association for Computational Linguistics},
  month     = oct,
  year      = {2020},
  doi       = {10.18653/v1/2020.emnlp-demos.6},
  eprint    = {1910.03771},
  archivePrefix = {arXiv}
}

@misc{ye2025dream,
  title={Dream 7B: Diffusion Large Language Models},
  author={Ye, Jiacheng and Xie, Zhihui and Zheng, Lin and Gao, Jiahui and Wu, Zirui and Jiang, Xin and Li, Zhenguo and Kong, Lingpeng},
  year={2025},
  eprint={2508.15487},
  archivePrefix={arXiv},
  url={https://arxiv.org/abs/2508.15487},
}

@misc{crawshaw2025exploration,
  title={An Exploration of Non-{E}uclidean Gradient Descent: {M}uon and its Many Variants},
  author={Crawshaw, Michael and Modi, Chirag and Liu, Mingrui and Gower, Robert M.},
  year={2025},
  eprint={2510.09827},
  archivePrefix={arXiv},
  url={https://arxiv.org/abs/2510.09827},
}

@inproceedings{schaipp2024momo,
  title={MoMo: Momentum Models for Adaptive Learning Rates},
  author={Fabian Schaipp and Ruben Ohana and Michael Eickenberg and Aaron Defazio and Robert M. Gower},
  editor={Ruslan Salakhutdinov and Zico Kolter and Katherine A. Heller and Adrian Weller and Nuria Oliver and Jonathan Scarlett and Felix Berkenkamp},
  booktitle={Forty-first International Conference on Machine Learning, {ICML} 2024, Vienna, Austria, July 21-27, 2024},
  series={Proceedings of Machine Learning Research},
  volume={235},
  pages={43542--43570},
  publisher={{PMLR} / OpenReview.net},
  year={2024},
  url={https://proceedings.mlr.press/v235/schaipp24a.html},
}

@inproceedings{chen2024lionk,
  title     = {Lion Secretly Solves a Constrained Optimization: As {L}yapunov Predicts},
  author    = {Chen, Lizhang and Liu, Bo and Liang, Kaizhao and Liu, Qiang},
  booktitle = {International Conference on Learning Representations},
  year      = {2024},
}

@inproceedings{defazio2019ineffectiveness,
  title={On the Ineffectiveness of Variance Reduced Optimization for Deep Learning},
  author={Defazio, Aaron and Bottou, L\'{e}on},
  booktitle={Advances in Neural Information Processing Systems},
  year={2019},
}

@misc{defazio2026schedulefreeplus,
  title={{ScheduleFree+}: Scaling Learning-Rate-Free \& Schedule-Free Learning to Large Language Models},
  author={Defazio, Aaron},
  year={2026},
  eprint={2605.19095},
  archivePrefix={arXiv},
  url={https://arxiv.org/abs/2605.19095},
}

@inproceedings{zhang2025criticalbatch,
  title={How Does Critical Batch Size Scale in Pre-training?},
  author={Zhang, Hanlin and Morwani, Depen and Vyas, Nikhil and Wu, Jingfeng and Zou, Difan and Ghai, Udaya and Foster, Dean and Kakade, Sham},
  booktitle={International Conference on Learning Representations},
  year={2025},
}

@misc{tyurin2026reversefisher,
  title={Global Convergence of Gradient Descent for Score Matching in {G}aussian Mixtures via Reverse {F}isher Divergence},
  author={Tyurin, Alexander},
  year={2026},
  eprint={2606.19876},
  archivePrefix={arXiv},
  url={https://arxiv.org/abs/2606.19876},
}

@misc{yang2022mutransfer,
  title        = {Tensor Programs {V}: Tuning Large Neural Networks via Zero-Shot Hyperparameter Transfer},
  author       = {Yang, Greg and Hu, Edward J. and Babuschkin, Igor and Sidor, Szymon and Liu, Xiaodong and Farhi, David and Ryder, Nick and Pachocki, Jakub and Chen, Weizhu and Gao, Jianfeng},
  year         = {2022},
  eprint       = {2203.03466},
  archivePrefix= {arXiv},
  url          = {https://arxiv.org/abs/2203.03466}
}

@article{nie2025llada,
  title   = {Large Language Diffusion Models},
  author  = {Nie, Shen and Zhu, Fengqi and You, Zebin and Zhang, Xiaolu and Ou, Jingyang and Hu, Jun and Zhou, Jun and Lin, Yankai and Wen, Ji-Rong and Li, Chongxuan},
  journal = {arXiv preprint arXiv:2502.09992},
  year    = {2025},
}

@article{lee2025genmol,
  title   = {{GenMol}: A Drug Discovery Generalist with Discrete Diffusion},
  author  = {Lee, Seul and Kreis, Karsten and Veccham, Srimukh Prasad and Liu, Meng and Reidenbach, Danny and Peng, Yuxing and Paliwal, Saee and Nie, Weili and Vahdat, Arash},
  journal = {arXiv preprint arXiv:2501.06158},
  year    = {2025},
}

@inproceedings{martens2015kfac,
  title     = {Optimizing Neural Networks with {K}ronecker-factored Approximate Curvature},
  author    = {Martens, James and Grosse, Roger},
  booktitle = {International Conference on Machine Learning},
  year      = {2015},
}

@misc{rdkit,
  title        = {{RDKit}: Open-Source Cheminformatics},
  author       = {{RDKit Project}},
  howpublished = {\url{https://www.rdkit.org}},
  year         = {2026},
  note         = {Software, version 2026.03.5},
}

@article{preuer2018fcd,
  title        = {{Fr\'echet ChemNet Distance}: A Metric for Generative Models for Molecules in Drug Discovery},
  author       = {Preuer, Kristina and Renz, Philipp and Unterthiner, Thomas and Hochreiter, Sepp and Klambauer, G{\"u}nter},
  journal      = {Journal of Chemical Information and Modeling},
  volume       = {58},
  number       = {9},
  pages        = {1736--1741},
  year         = {2018},
  doi          = {10.1021/acs.jcim.8b00234},
  eprint       = {1803.09518},
  archivePrefix= {arXiv}
}

@misc{zhang2020clipping,
      title={Why gradient clipping accelerates training: A theoretical justification for adaptivity}, 
      author={Jingzhao Zhang and Tianxing He and Suvrit Sra and Ali Jadbabaie},
      year={2020},
      eprint={1905.11881},
      archivePrefix={arXiv},
      primaryClass={math.OC},
      url={https://arxiv.org/abs/1905.11881}, 
}

@misc{li2023generalized,
      title={Convex and Non-convex Optimization Under Generalized Smoothness}, 
      author={Haochuan Li and Jian Qian and Yi Tian and Alexander Rakhlin and Ali Jadbabaie},
      year={2023},
      eprint={2306.01264},
      archivePrefix={arXiv},
      primaryClass={math.OC},
      url={https://arxiv.org/abs/2306.01264}, 
}

@misc{koloskova2023clipping,
      title={Revisiting Gradient Clipping: Stochastic bias and tight convergence guarantees}, 
      author={Anastasia Koloskova and Hadrien Hendrikx and Sebastian U. Stich},
      year={2023},
      eprint={2305.01588},
      archivePrefix={arXiv},
      primaryClass={cs.LG},
      url={https://arxiv.org/abs/2305.01588}, 
}

@misc{crawshaw2022signsgd,
      title={Robustness to Unbounded Smoothness of Generalized SignSGD}, 
      author={Michael Crawshaw and Mingrui Liu and Francesco Orabona and Wei Zhang and Zhenxun Zhuang},
      year={2022},
      eprint={2208.11195},
      archivePrefix={arXiv},
      primaryClass={cs.LG},
      url={https://arxiv.org/abs/2208.11195}, 
}

@misc{gorbunov2024l0l1,
      title={Methods for Convex $(L_0,L_1)$-Smooth Optimization: Clipping, Acceleration, and Adaptivity}, 
      author={Eduard Gorbunov and Nazarii Tupitsa and Sayantan Choudhury and Alen Aliev and Peter Richt\'arik and Samuel Horv\'ath and Martin Tak\'a\v{c}},
      year={2024},
      eprint={2409.14989},
      archivePrefix={arXiv},
      primaryClass={math.OC},
      url={https://arxiv.org/abs/2409.14989}, 
}

\newpage
\appendix
\section{Implementations}
\label{app:optimizers}

Four of the seven are used through their reference implementations: AdamW from
PyTorch \citep{paszke2019pytorch}, Lion from \texttt{lion-pytorch}
\citep{wang2023lionpytorch}, SOAP from \texttt{pytorch-optimizer}
\citep{kim2021pytorchoptimizer},
and Schedule-Free from the authors' \texttt{schedulefree} package. Muon, MARS
and MARS-M are ported from the equations in the original papers, since no
reference package covers the variants we need. Pseudocode for all seven is given
in Section~\ref{app:pseudocode}.

Both MARS ports use the approximate gradient correction, which reuses the
previous step's gradient $g_{t-1}$ instead of recomputing the previous
parameters' gradient on the current batch; the exact form needs a second
backward pass per step, and \citet{yuan2025mars} recommend the approximation.
Neither port applies the norm clip on $c_t$ of the published algorithms, and
the pseudocode below omits it.

Muon and MARS-M run their AdamW branch with a separate learning rate
\texttt{adamw\_lr}, tuned as its own axis, following the Moonlight variant
\citep{liu2025moonlight}. Schedule-Free runs with a constant learning rate
after warmup, while the other six use warmup-stable-decay.

\section{Pseudocode}
\label{app:pseudocode}

Each algorithm below is one full training loop, as run. Throughout, $\theta_t$ are the parameters at step $t$, $\xi_t$ is
the minibatch drawn at step $t$, $g_t = \nabla f(\theta_{t-1}, \xi_t)$ is the
stochastic gradient, $\eta_t$ is the learning rate after the schedule multiplier
is applied, and $\lambda$ is the decoupled weight decay. All products, squares,
divisions and comparisons between tensors of the same shape are element-wise.
$\odot$ denotes the element-wise product and $\lVert\cdot\rVert_F$ the Frobenius
norm.

\begin{algorithm}[htbp]
\caption{AdamW \citep{loshchilov2019adamw}. The weight decay on line~\ref{ln:awwd}
acts on $\theta$ directly, outside the preconditioner, unlike an $\ell_2$
penalty folded into $g_t$.}
\label{alg:adamw}
\begin{algorithmic}[1]
\Require learning rate $\eta$, betas $\beta_1, \beta_2 \in [0,1)$, $\epsilon$,
weight decay $\lambda$, steps $T$
\State $m_0 \gets 0$, \; $v_0 \gets 0$
\For{$t = 1$ \textbf{to} $T$}
  \State $g_t \gets \nabla f(\theta_{t-1}, \xi_t)$
  \State $m_t \gets \beta_1 m_{t-1} + (1-\beta_1)\, g_t$
    \Comment{first moment}
  \State $v_t \gets \beta_2 v_{t-1} + (1-\beta_2)\, g_t^2$
    \Comment{second moment}
  \State $\hat{m}_t \gets m_t / (1-\beta_1^{\,t})$, \;
         $\hat{v}_t \gets v_t / (1-\beta_2^{\,t})$
    \Comment{bias correction}
  \State $\theta_t \gets \theta_{t-1}
         - \eta_t \big( \hat{m}_t / (\sqrt{\hat{v}_t} + \epsilon)
         + \lambda\, \theta_{t-1} \big)$ \label{ln:awwd}
\EndFor
\State \Return $\theta_T$
\end{algorithmic}
\end{algorithm}

\begin{algorithm}[htbp]
\caption{Lion \citep{chen2023lion}. The update is the sign of an interpolation
that uses $\beta_1$, while the stored momentum is a separate interpolation that
uses $\beta_2$ and is written after the update is formed. Because $\operatorname{sign}$
gives every coordinate unit magnitude, Lion needs a smaller $\eta$ and a larger
$\lambda$ than AdamW to reach the same effective step and decay.}
\label{alg:lion}
\begin{algorithmic}[1]
\Require learning rate $\eta$, betas $\beta_1, \beta_2$ (tuned $0.9$,
$0.99$), weight decay $\lambda$, steps $T$
\State $m_0 \gets 0$
\For{$t = 1$ \textbf{to} $T$}
  \State $g_t \gets \nabla f(\theta_{t-1}, \xi_t)$
  \State $c_t \gets \beta_1 m_{t-1} + (1-\beta_1)\, g_t$
    \Comment{interpolate for the update}
  \State $u_t \gets \operatorname{sign}(c_t)$
  \State $\theta_t \gets \theta_{t-1} - \eta_t \big( u_t + \lambda\,\theta_{t-1} \big)$
  \State $m_t \gets \beta_2 m_{t-1} + (1-\beta_2)\, g_t$
    \Comment{momentum updated after the step}
\EndFor
\State \Return $\theta_T$
\end{algorithmic}
\end{algorithm}

\begin{algorithm}[htbp]
\caption{Newton--Schulz orthogonalization, $\textsc{NS}_5$
\citep{bjorck1971orthogonal,higham2008functions}. Five iterations of a quintic
polynomial drive the singular values of $G$ towards one, so the output
approximates the orthogonal factor $UV^{\!\top}$ of $G = U\Sigma V^{\!\top}$
without an SVD. The coefficients are tuned so the iteration converges quickly
for small initial singular values, and it runs in bfloat16.}
\label{alg:ns}
\begin{algorithmic}[1]
\Require $G \in \mathbb{R}^{m \times n}$, steps $N = 5$,
$(a,b,c) = (3.4445, -4.7750, 2.0315)$
\State $X \gets G / (\lVert G \rVert_F + 10^{-7})$
  \Comment{scale so the spectrum lies in the basin}
\If{$m > n$}
  \State $X \gets X^{\!\top}$ \Comment{operate on the wide orientation}
\EndIf
\For{$i = 1$ \textbf{to} $N$}
  \State $A \gets X X^{\!\top}$
  \State $B \gets b\,A + c\,A A$
  \State $X \gets a\,X + B X$
    \Comment{$X \mapsto aX + bXX^{\!\top}X + c(XX^{\!\top})^2 X$}
\EndFor
\If{$m > n$}
  \State $X \gets X^{\!\top}$ \Comment{undo the transpose}
\EndIf
\State \Return $X \approx UV^{\!\top}$
\end{algorithmic}
\end{algorithm}

\begin{algorithm}[htbp]
\caption{Muon \citep{jordan2024muon} with the scaling and decay of
\citet{liu2025moonlight}. Only hidden matrices are orthogonalized; scalars,
embeddings and the output head take an AdamW step with their own rate
\texttt{adamw\_lr}, since orthogonalization is meaningless for them. The factor
$0.2\sqrt{\max(m,n)}$ on line~\ref{ln:muonrms} matches the update RMS to
AdamW's, which is what lets a rate tuned for AdamW transfer.}
\label{alg:muon}
\begin{algorithmic}[1]
\Require rates $\eta$ and $\eta_{\text{aw}}$, momentum $\mu$, weight decay
$\lambda$, \texttt{nesterov} flag, steps $T$
\State partition parameters into matrices $\mathcal{M}$ and the rest
$\mathcal{A}$; \; $B_0 \gets 0$
\For{$t = 1$ \textbf{to} $T$}
  \For{each $\theta \in \mathcal{A}$}
    \State AdamW step (Algorithm~\ref{alg:adamw}) with rate $\eta_{\text{aw}}$
  \EndFor
  \For{each $\theta \in \mathcal{M}$, of shape $m \times n$}
    \State $G_t \gets \nabla_\theta f(\theta_{t-1}, \xi_t)$
    \State $B_t \gets \mu B_{t-1} + G_t$ \Comment{heavy-ball buffer}
    \State $U_t \gets G_t + \mu B_t$ \textbf{if} \texttt{nesterov} \textbf{else} $B_t$
      \Comment{Nesterov form \citep{sutskever2013momentum}}
    \State $O_t \gets \textsc{NS}_5(U_t)$ \Comment{Algorithm~\ref{alg:ns}}
    \State $\eta'_t \gets 0.2\, \eta_t \sqrt{\max(m,n)}$ \label{ln:muonrms}
    \State $\theta_t \gets (1 - \eta_t \lambda)\, \theta_{t-1} - \eta'_t\, O_t$
      \Comment{decoupled decay, then the step}
  \EndFor
\EndFor
\State \Return $\theta_T$
\end{algorithmic}
\end{algorithm}

\begin{algorithm}[htbp]
\caption{SOAP \citep{vyas2025soap}. Adam is run inside the eigenbasis of
Shampoo's \citep{gupta2018shampoo} two Kronecker factors. The momentum is kept
in the original basis and rotated each step, while the second moment is
accumulated in the rotated basis. Refreshing $Q_L, Q_R$ only every $f$ steps is
what makes the method affordable; one power iteration plus a QR is enough
because the previous basis is a warm start. Fixing both $Q$ to the identity
recovers Adam.}
\label{alg:soap}
\begin{algorithmic}[1]
\Require rate $\eta$, betas $\beta_1,\beta_2$, $\epsilon$, decay $\lambda$,
precondition frequency $f$, steps $T$
\State $L_0, R_0, M_0, V_0 \gets 0$; \; $Q_L, Q_R \gets$ eigenvectors of
$L_0, R_0$
\For{$t = 1$ \textbf{to} $T$}
  \State $G_t \gets \nabla f(\theta_{t-1}, \xi_t)$ \Comment{for a matrix layer $m \times n$}
  \State $\tilde{G}_t \gets Q_L^{\!\top} G_t\, Q_R$ \Comment{rotate the gradient}
  \State $M_t \gets \beta_1 M_{t-1} + (1-\beta_1) G_t$
    \Comment{momentum in the original basis}
  \State $\tilde{M}_t \gets Q_L^{\!\top} M_t\, Q_R$
  \State $V_t \gets \beta_2 V_{t-1} + (1-\beta_2)\, \tilde{G}_t \odot \tilde{G}_t$
    \Comment{second moment in the rotated basis}
  \State $\tilde{N}_t \gets \tilde{M}_t / (\sqrt{V_t} + \epsilon)$
    \Comment{the Adam step, taken in the eigenbasis}
  \State $N_t \gets Q_L \tilde{N}_t\, Q_R^{\!\top}$ \Comment{rotate back}
  \State $\theta_t \gets \theta_{t-1} - \eta_t \big( N_t + \lambda\, \theta_{t-1} \big)$
  \State $L_t \gets \beta_2 L_{t-1} + (1-\beta_2)\, G_t G_t^{\!\top}$
  \State $R_t \gets \beta_2 R_{t-1} + (1-\beta_2)\, G_t^{\!\top} G_t$
  \If{$t \bmod f = 0$}
    \State $Q_L \gets \textsc{QR}(L_t Q_L)$, \;
           $Q_R \gets \textsc{QR}(R_t Q_R)$
      \Comment{one power iteration, warm-started}
  \EndIf
\EndFor
\State \Return $\theta_T$
\end{algorithmic}
\end{algorithm}

\begin{algorithm}[htbp]
\caption{MARS \citep{yuan2025mars}, AdamW instantiation, approximate variant.
Line~\ref{ln:marsc} extrapolates along the gradient difference, which is the
STORM \citep{cutkosky2019storm} correction scaled by $\gamma$; $\gamma = 0$
recovers AdamW exactly. Both moments are built from the corrected $c_t$ rather
than from $g_t$, which the authors note is what keeps the per-coordinate scale
right. We use the approximate form, which reuses the previous step's gradient
instead of recomputing $\nabla f(\theta_{t-2}, \xi_t)$ and so costs one backward
pass per step. Our port omits the norm clip on $c_t$.}
\label{alg:mars}
\begin{algorithmic}[1]
\Require rate $\eta$, betas $\beta_1,\beta_2$, $\epsilon$, decay $\lambda$,
scale $\gamma$, steps $T$
\State $m_0, v_0 \gets 0$; \; $g_0 \gets 0$
\For{$t = 1$ \textbf{to} $T$}
  \State $g_t \gets \nabla f(\theta_{t-1}, \xi_t)$
  \State $c_t \gets g_t + \gamma \dfrac{\beta_1}{1-\beta_1}\,(g_t - g_{t-1})$
    \label{ln:marsc} \Comment{scaled gradient correction}
  \State $m_t \gets \beta_1 m_{t-1} + (1-\beta_1)\, c_t$
  \State $v_t \gets \beta_2 v_{t-1} + (1-\beta_2)\, c_t^2$
    \Comment{note $c_t$, not $g_t$}
  \State $\hat{m}_t \gets m_t / (1-\beta_1^{\,t})$, \;
         $\hat{v}_t \gets v_t / (1-\beta_2^{\,t})$
  \State $\theta_t \gets \theta_{t-1}
         - \eta_t \big( \hat{m}_t / (\sqrt{\hat{v}_t} + \epsilon)
         + \lambda\, \theta_{t-1} \big)$
\EndFor
\State \Return $\theta_T$
\end{algorithmic}
\end{algorithm}

\begin{algorithm}[htbp]
\caption{MARS-M \citep{liu2025marsm}. The same corrected gradient as
Algorithm~\ref{alg:mars} is fed into Muon's orthogonalized step instead of
AdamW's, so the difference between this and Algorithm~\ref{alg:muon} is exactly
the correction on line~\ref{ln:marsmc}. Note the momentum here is a convex
combination, $(1-\mu)$ weighting $c_t$, where plain Muon adds $G_t$ undamped.
Scalars, embeddings and the output head take the AdamW branch as in Muon.}
\label{alg:marsm}
\begin{algorithmic}[1]
\Require rates $\eta$ and $\eta_{\text{aw}}$, momentum $\mu$, decay $\lambda$,
scale $\gamma$, steps $T$
\State $M_0 \gets 0$, \; $G_0 \gets 0$
\For{$t = 1$ \textbf{to} $T$}
  \For{each matrix $\theta \in \mathcal{M}$ of shape $m \times n$}
    \State $G_t \gets \nabla_\theta f(\theta_{t-1}, \xi_t)$
    \State $C_t \gets G_t + \gamma \dfrac{\mu}{1-\mu}\,(G_t - G_{t-1})$
      \label{ln:marsmc}
    \State $M_t \gets \mu M_{t-1} + (1-\mu)\, C_t$
    \State $O_t \gets \textsc{NS}_5(M_t)$
    \State $\eta'_t \gets 0.2\, \eta_t \sqrt{\max(m,n)}$
    \State $\theta_t \gets (1 - \eta_t \lambda)\, \theta_{t-1} - \eta'_t\, O_t$
  \EndFor
  \State remaining parameters: AdamW step with rate $\eta_{\text{aw}}$
\EndFor
\State \Return $\theta_T$
\end{algorithmic}
\end{algorithm}

\begin{algorithm}[htbp]
\caption{Schedule-Free AdamW \citep{defazio2024schedulefree}. Three sequences
are kept: gradients are taken at $y_t$, the Adam step is applied to the base
iterate $z_t$, and the parameters that are evaluated are the running average
$x_t$. There is no first-moment buffer, so memory matches AdamW; momentum comes
entirely from the interpolation on line~\ref{ln:sfy}. The averaging weights are
proportional to $\eta_t^2$, which matters while the warmup rate is still rising,
and the decay is applied at $y_t$.}
\label{alg:sf}
\begin{algorithmic}[1]
\Require rate $\eta$, $\beta_1$ (interpolation), $\beta_2$, $\epsilon$, decay
$\lambda$, warmup $T_{\text{w}}$, steps $T$
\State $z_1 \gets \theta_1$, \; $x_1 \gets \theta_1$, \; $v_0 \gets 0$
\For{$t = 1$ \textbf{to} $T$}
  \State $y_t \gets (1-\beta_1)\, z_t + \beta_1\, x_t$ \label{ln:sfy}
    \Comment{momentum by interpolation}
  \State $g_t \gets \nabla f(y_t, \xi_t)$
    \Comment{gradient at $y$, not at $x$ or $z$}
  \State $v_t \gets \beta_2 v_{t-1} + (1-\beta_2)\, g_t^2$
  \State $\eta_t \gets \eta \sqrt{1-\beta_2^{\,t}} \cdot \min(1,\, t/T_{\text{w}})$
    \Comment{warmup and bias correction fused}
  \State $z_{t+1} \gets z_t - \eta_t\, g_t / (\sqrt{v_t} + \epsilon)
         - \eta_t \lambda\, y_t$
  \State $c_{t+1} \gets \eta_t^2 \big/ \sum_{i \le t} \eta_i^2$
    \Comment{weighted, not uniform}
  \State $x_{t+1} \gets (1-c_{t+1})\, x_t + c_{t+1}\, z_{t+1}$
\EndFor
\State \Return $x_{T+1}$ \Comment{evaluate and checkpoint at $x$}
\end{algorithmic}
\end{algorithm}

\section{How our budgets compare to published ones}
\label{app:budgets}

Two of our four tasks are deliberately truncated, one is a full-budget
replication, and CelebA-64 sits in between at one fifth of the DDIM paper's
training length.

QM9 is a full-budget replication. Published UDLM results use 25k steps at
batch 2048 with context 32 and a 92.4M parameter transformer
\citep{schiff2025guidance}; we use 25k steps, batch 2048, context 32 and
90.2M parameters. Our perplexities land in the published band, so the QM9 column can
be read against the literature directly.

text8 is truncated by a factor of 33. Published text8 results
\citep{austin2021d3pm,lou2024sedd,shi2024md4} use 70M to 92M
parameter 12-layer transformers trained for 1M steps at batch 512 and context
256, about 131B characters or 1450 epochs over the 90M-character training
split. We train a 16.6M parameter 8-layer model for 30k steps at the same
batch size and context, 3.9B characters or 44 epochs: 33 times fewer steps at
4 to 5.6 times fewer parameters. Our bits per character are therefore far
from the published 1.37 to 1.45 and are only compared within our own table.

LM1B is truncated the same way: published results use 1M steps and we use 30k,
with the same architecture and per-step token count.

For CelebA-64 the headline FID is computed with torch-fidelity
\citep{obukhov2020torchfidelity}, whose Inception features follow
\citet{szegedy2016inception}, from 50k samples at 100 DDIM steps
\citep{song2021ddim} against the training-set statistics, using EMA weights.
The DDIM paper reports 6.53 at 100 steps for a model trained five times longer
on a different crop, so the scales are close but not identical, and we use
ours only to rank optimizers against each other.

\section{Architecture details}
\label{app:arch}

\begin{table}[htbp]
\caption{The four training tasks. Steps is the full benchmark budget, tune
steps is the reduced budget used during hyperparameter search.}
\label{tab:tasks}
\centering
\small
\begin{tabular}{lrlrrrr}
\toprule
task & params & shape & batch & ctx & steps & tune steps \\
\midrule
text8 (MDLM) & 16.6M & 8L/256d/8h & 512 & 256 & 30000 & 6000 \\
QM9 (UDLM) & 90.2M & 12L/768d/12h & 2048 & 32 & 25000 & 5000 \\
LM1B (DUO) & 137.1M & 12L/768d/12h & 512 & 128 & 30000 & 6000 \\
CelebA-64 (DDPM) & 78.7M & UNet, ch 128 & 128 & 64px & 100000 & 20000 \\
\bottomrule
\end{tabular}

\end{table}
 The three text and molecule tasks share one backbone,
a pre-norm transformer with bidirectional self-attention, since every position
may condition on every other. Token embeddings are added to a learned position
embedding. The diffusion time $t \in [0,1]$ is mapped through a sinusoidal
embedding and a two-layer MLP. It enters every block through adaptive layer
norm, as in DiT: the layer norms carry no affine parameters of their own, and
a zero-initialized linear layer predicts their scale and shift from the time
embedding. At initialization the network is therefore a plain transformer. Attention uses
$d_{\text{model}}/n_{\text{head}}$ dimensions per head and the MLP widens by
four with GELU. The output head is untied from the input embedding. Sizes are
8 layers, width 256, 8 heads for text8 (16.6M parameters) and 12 layers, width
768, 12 heads for QM9 (90.2M) and LM1B (137.1M, the difference being the
30k-token vocabulary).

CelebA uses the U-Net of DDPM as implemented in \texttt{diffusers}
\citep{vonplaten2022diffusers}: five resolution levels with channel widths
$128 \times (1, 2, 2, 2, 4)$, two residual blocks per level, self-attention at
the $16\times16$ level, and sinusoidal timestep embeddings, for 78.7M
parameters. Sampling for FID uses an exponential moving average of the weights
with decay 0.9999.

\section{Data preparation}
\label{app:data}

\textbf{text8.} The raw 100M-character file is mapped to integer ids, 26 letters
plus space, with one extra id reserved for the MDLM mask symbol. The first 90M
characters are the training split, the next 5M validation and the last 5M test,
which is the split used by every text8 result we compare against. Training
windows of 256 characters are drawn at uniformly random offsets, so each epoch
sees a different segmentation.

\textbf{QM9.} We take the canonical SMILES field of the \texttt{yairschiff/qm9}
release used by UDLM and tokenize with the matching 40-symbol tokenizer, which
treats multi-character atoms and bracketed groups as single tokens. Sequences
are padded to 32 with a dedicated pad id that is masked out of the loss and the
evaluation. A fixed random 5\%/5\% of molecules is held out for validation and
test.

\textbf{LM1B.} We apply MDLM's detokenizer to each sentence, then the
\texttt{bert-base-uncased} WordPiece tokenizer, and pad or truncate to 128
tokens. Unlike MDLM, which trains on the full 30M sentences for a million
steps, we stream the first two million sentences; at 30k steps and batch 512 the
model makes roughly eight passes over this subset.

\textbf{CelebA.} Images come from the aligned-and-cropped release at
$178\times218$. We take a central $140\times140$ crop, resize to $64\times64$,
and scale pixels to $[-1,1]$. Training applies a random horizontal flip; no
other augmentation is used. The crop matches NCSN \citep{song2019ncsn} and
the DDPM line. The DDIM paper instead uses a $128$ crop taken off-center, which
is one reason our absolute FID is not comparable to theirs. Reference statistics
for FID are computed over 50k training-split images.

\section{Search grids and selected values}
\label{app:grids}

Three axes are shared. \texttt{lr} takes one of $\{10^{-5}, 3\cdot10^{-5},
10^{-4}, 3\cdot10^{-4}, 10^{-3}, 3\cdot10^{-3}, 10^{-2}, 3\cdot10^{-2}\}$,
\texttt{weight\_decay} one of $\{0, 0.01, 0.1, 0.3, 1, 3\}$, \texttt{grad\_clip}
one of $\{0, 1\}$. Six studies departed from these, listed in
Table~\ref{tab:griddev}: MARS was given a narrower \texttt{lr} range after
early trials diverged at the extremes, Schedule-Free was given one
extra \texttt{weight\_decay} value on text8, and two CelebA-64 studies ran
before the shared axes were finalised. The
optimizer-specific axes are in
Table~\ref{tab:grid}. Table~\ref{tab:winners} lists the configuration selected
for every task and optimizer, which is exactly what the benchmark runs used.

\begin{table}[t]
\caption{Optimizer-specific search axes, the union of the value sets
searched across studies; a few studies searched narrower sets on some axes
(MARS's smallest \texttt{gamma} on text8 only, narrower \texttt{adamw\_lr}
and \texttt{momentum} in the early CelebA-64 studies), and the edge
selections in the text are relative to the set each study searched. Every
optimizer additionally sweeps the three shared axes listed in the text.}
\label{tab:grid}
\centering
\small
\resizebox{\textwidth}{!}{\begin{tabular}{llp{7.6cm}}
\toprule
optimizer & axis & values \\
\midrule
AdamW & \texttt{beta1} & $\{$$0.5$, $0.8$, $0.9$, $0.95$, $0.99$, $0.999$$\}$ \\
 & \texttt{beta2} & $\{$$0.5$, $0.8$, $0.9$, $0.95$, $0.99$, $0.999$$\}$ \\
Lion & \texttt{beta1} & $\{$$0.5$, $0.8$, $0.9$, $0.95$, $0.99$, $0.999$$\}$ \\
 & \texttt{beta2} & $\{$$0.5$, $0.8$, $0.9$, $0.95$, $0.99$, $0.999$$\}$ \\
Muon & \texttt{adamw\_lr} & $\{$$3\!\cdot\!10^{-6}$, $1\!\cdot\!10^{-5}$, $3\!\cdot\!10^{-5}$, $1\!\cdot\!10^{-4}$, $3\!\cdot\!10^{-4}$, $1\!\cdot\!10^{-3}$, $3\!\cdot\!10^{-3}$, $0.01$, $0.03$$\}$ \\
 & \texttt{momentum} & $\{$$0$, $0.2$, $0.5$, $0.8$, $0.9$, $0.95$, $0.99$, $0.999$$\}$ \\
 & \texttt{nesterov} & $\{$$0$, $1$$\}$ \\
Schedule-Free & \texttt{beta1} & $\{$$0.5$, $0.8$, $0.9$, $0.95$, $0.99$, $0.999$$\}$ \\
 & \texttt{beta2} & $\{$$0.5$, $0.8$, $0.9$, $0.95$, $0.99$, $0.999$$\}$ \\
SOAP & \texttt{beta1} & $\{$$0.5$, $0.8$, $0.9$, $0.95$, $0.99$, $0.999$$\}$ \\
 & \texttt{beta2} & $\{$$0.5$, $0.8$, $0.9$, $0.95$, $0.99$, $0.999$$\}$ \\
 & \texttt{precondition\_frequency} & $\{$$10$, $50$$\}$ \\
MARS & \texttt{beta1} & $\{$$0.5$, $0.8$, $0.9$, $0.95$, $0.99$, $0.999$$\}$ \\
 & \texttt{beta2} & $\{$$0.5$, $0.8$, $0.9$, $0.95$, $0.99$, $0.999$, $0.9999$$\}$ \\
 & \texttt{gamma} & $\{$$3\!\cdot\!10^{-3}$, $0.01$, $0.025$, $0.075$, $0.1$, $0.5$, $1$$\}$ \\
MARS-M & \texttt{adamw\_lr} & $\{$$3\!\cdot\!10^{-6}$, $1\!\cdot\!10^{-5}$, $3\!\cdot\!10^{-5}$, $1\!\cdot\!10^{-4}$, $3\!\cdot\!10^{-4}$, $1\!\cdot\!10^{-3}$, $3\!\cdot\!10^{-3}$, $0.01$, $0.03$$\}$ \\
 & \texttt{gamma} & $\{$$0.01$, $0.025$, $0.075$, $0.1$, $0.5$, $1$$\}$ \\
 & \texttt{momentum} & $\{$$0.5$, $0.8$, $0.9$, $0.95$, $0.99$, $0.999$$\}$ \\
\bottomrule
\end{tabular}
}
\end{table}

\begin{table}[htbp]
\caption{Every study whose shared axes differed from the common grid, generated
from the values each study actually searched.}
\label{tab:griddev}
\centering
\small
\begin{tabular}{lllll}
\toprule
task & optimizer & axis & searched & common \\
\midrule
text8 & Schedule-Free & weight\_decay & $\{0, \ldots, 10\}$ (7) & (6) \\
text8 & MARS & lr & $\{0.0001, \ldots, 0.01\}$ (5) & (8) \\
qm9 & MARS & lr & $\{0.0001, \ldots, 0.01\}$ (5) & (8) \\
celeba64 & Muon & lr & $\{1e-05, \ldots, 0.01\}$ (7) & (8) \\
celeba64 & MARS & lr & $\{0.0001, \ldots, 0.01\}$ (5) & (8) \\
celeba64 & MARS-M & lr & $\{3e-05, \ldots, 0.003\}$ (5) & (8) \\
\bottomrule
\end{tabular}

\end{table}

\paragraph{Selections on a grid edge.} Ten remain. On text8, Lion takes the
largest \texttt{weight\_decay} and MARS the smallest \texttt{gamma}. On QM9,
MARS-M takes the smallest \texttt{momentum}. On LM1B, AdamW and SOAP take
the largest \texttt{beta2} and Schedule-Free the smallest \texttt{beta1}. On
CelebA-64, Muon takes the
smallest \texttt{adamw\_lr}, and MARS-M the largest \texttt{adamw\_lr}, the
smallest \texttt{lr} and the smallest \texttt{gamma}. Three of the four CelebA-64 cases (Muon's \texttt{adamw\_lr}, and MARS-M's
\texttt{adamw\_lr} and \texttt{lr}) sit on axes whose widened grids were never
searched. For these the reported value is a lower bound on what tuning could
reach, so any comparison involving them understates that optimizer.

\begin{table}[htbp]
\caption{Selected hyperparameters. These are the configurations retrained at the
full budget with three seeds.}
\label{tab:winners}
\centering
\small
\resizebox{\textwidth}{!}{\begin{tabular}{llrrrl}
\toprule
task & optimizer & lr & wd & clip & other \\
\midrule
text8 (MDLM) & AdamW & 0.001 & 0.3 & 0 & \scriptsize \texttt{beta1=0.8}, \texttt{beta2=0.8} \\
 & Lion & 0.0003 & 3 & 0 & \scriptsize \texttt{beta1=0.8}, \texttt{beta2=0.9} \\
 & Muon & 0.01 & 0.1 & 1 & \scriptsize \texttt{adamw\_lr=0.001}, \texttt{momentum=0.5}, \texttt{nesterov=False} \\
 & Schedule-Free & 0.01 & 1 & 0 & \scriptsize \texttt{beta1=0.95}, \texttt{beta2=0.8} \\
 & SOAP & 0.003 & 0.3 & 0 & \scriptsize \texttt{precondition\_frequency=10}, \texttt{beta1=0.95}, \texttt{beta2=0.95} \\
 & MARS & 0.001 & 1 & 0 & \scriptsize \texttt{gamma=0.003}, \texttt{beta1=0.9}, \texttt{beta2=0.9} \\
 & MARS-M & 0.003 & 0.3 & 1 & \scriptsize \texttt{adamw\_lr=0.0003}, \texttt{momentum=0.95}, \texttt{gamma=0.025} \\
\midrule
QM9 (UDLM) & AdamW & 0.0003 & 0.01 & 1 & \scriptsize \texttt{beta1=0.8}, \texttt{beta2=0.9} \\
 & Lion & 0.0001 & 0.3 & 1 & \scriptsize \texttt{beta1=0.8}, \texttt{beta2=0.9} \\
 & Muon & 0.003 & 0.1 & 1 & \scriptsize \texttt{adamw\_lr=1e-05}, \texttt{momentum=0.5}, \texttt{nesterov=False} \\
 & Schedule-Free & 0.001 & 1 & 1 & \scriptsize \texttt{beta1=0.9}, \texttt{beta2=0.95} \\
 & SOAP & 0.001 & 0.3 & 0 & \scriptsize \texttt{precondition\_frequency=10}, \texttt{beta1=0.95}, \texttt{beta2=0.9} \\
 & MARS & 0.001 & 0.01 & 1 & \scriptsize \texttt{gamma=0.075}, \texttt{beta1=0.95}, \texttt{beta2=0.9} \\
 & MARS-M & 0.003 & 0.1 & 1 & \scriptsize \texttt{adamw\_lr=0.001}, \texttt{momentum=0.5}, \texttt{gamma=0.1} \\
\midrule
LM1B (DUO) & AdamW & 0.0003 & 0.3 & 1 & \scriptsize \texttt{beta1=0.8}, \texttt{beta2=0.999} \\
 & Lion & 0.0003 & 0.3 & 1 & \scriptsize \texttt{beta1=0.9}, \texttt{beta2=0.95} \\
 & Muon & 0.001 & 0 & 1 & \scriptsize \texttt{adamw\_lr=0.001}, \texttt{momentum=0.8}, \texttt{nesterov=True} \\
 & Schedule-Free & 0.001 & 0.3 & 1 & \scriptsize \texttt{beta1=0.5}, \texttt{beta2=0.99} \\
 & SOAP & 0.0003 & 1 & 1 & \scriptsize \texttt{precondition\_frequency=10}, \texttt{beta1=0.9}, \texttt{beta2=0.999} \\
 & MARS & 0.001 & 0.01 & 1 & \scriptsize \texttt{gamma=0.075}, \texttt{beta1=0.9}, \texttt{beta2=0.99} \\
 & MARS-M & 0.001 & 0.01 & 1 & \scriptsize \texttt{adamw\_lr=0.001}, \texttt{momentum=0.9}, \texttt{gamma=0.075} \\
\midrule
CelebA-64 (DDPM) & AdamW & 0.0003 & 0.01 & 0 & \scriptsize \texttt{beta1=0.95}, \texttt{beta2=0.99} \\
 & Lion & 0.0001 & 0.3 & 1 & \scriptsize \texttt{beta1=0.9}, \texttt{beta2=0.99} \\
 & Muon & 3e-05 & 0 & 0 & \scriptsize \texttt{adamw\_lr=3e-05}, \texttt{momentum=0.99}, \texttt{nesterov=True} \\
 & Schedule-Free & 0.003 & 0.01 & 1 & \scriptsize \texttt{beta1=0.95}, \texttt{beta2=0.95} \\
 & SOAP & 0.001 & 0.01 & 0 & \scriptsize \texttt{precondition\_frequency=10}, \texttt{beta1=0.9}, \texttt{beta2=0.95} \\
 & MARS & 0.001 & 0.01 & 1 & \scriptsize \texttt{gamma=0.1}, \texttt{beta1=0.9}, \texttt{beta2=0.8} \\
 & MARS-M & 3e-05 & 0 & 0 & \scriptsize \texttt{adamw\_lr=0.003}, \texttt{momentum=0.95}, \texttt{gamma=0.01} \\
\bottomrule
\end{tabular}
}
\end{table}

\section{Statistics}
\label{app:stats}

Comparisons against AdamW are paired within seed, which removes the shared
effect of the seed before the difference is taken \citep{student1908probable}.
We call a difference resolved when it exceeds twice the standard error of the
paired difference across the three seeds. We report each comparison on its own
and do not correct for the six simultaneous tests per task
\citep{holm1979sequential}: the largest resolved gaps clear the threshold
many times over (MARS on text8 at 59 standard errors of its paired
difference, Muon on LM1B at 26) and would survive any standard correction,
while the smallest, Lion on CelebA-64 at 2.2 and Muon on QM9 at 2.4, are
marginal calls. A correction can only demote differences, never rescue a tie.
Seed-to-seed variation is of the same order as several of the gaps we
report, which is why the pairing matters \citep{bouthillier2021variance}.

\section{Evaluation protocol}
\label{app:eval}

\textbf{Discrete tasks.} The continuous-time bound is estimated with 32 Monte
Carlo timestep draws over five fixed evaluation batches of 128 sequences, and
reported as bits per character on text8 and as per-token perplexity, two to
the power of the per-token bits, on QM9 and LM1B. The evaluation batches are
identical across optimizers and seeds. On text8 this is the validation split, characters 90M to 95M of the
cleaned dump. On QM9 and LM1B padding positions are excluded from both the
numerator and the denominator.

\textbf{CelebA-64.} The denoising loss is the $\epsilon$-prediction mean squared
error on held-out images with freshly drawn timesteps and noise at each
evaluation. The monitor FID is computed every 5000 steps from 10k samples at 50
DDIM steps \citep{song2021ddim} from the EMA weights \citep{song2020improved},
scored against the validation split. The headline FID is computed once at the
end from 50k samples at 100 DDIM steps from the same EMA weights, scored
against 50k training-split images, both with torch-fidelity
\citep{obukhov2020torchfidelity}. Preprocessing and resizing choices shift FID
on their own \citep{parmar2022cleanfid}, and the DDIM paper uses a $128$ pixel
crop taken off-center where we use the $140$ center crop, so our headline values
are close to that paper's scale but not on it. Both FIDs are internally
consistent and we use them only to compare optimizers.

\section{Cost measurement}
\label{app:cost}

\begin{table}[htbp]
\caption{Median time per training step and peak memory, both relative to AdamW
on the same task. Values above 1 are more expensive; the tint grows with the
overhead.}
\label{tab:cost}
\centering
\small
\begin{tabular}{lcccccc}
\toprule
 & \multicolumn{2}{c}{text8 (MDLM)} & \multicolumn{2}{c}{QM9 (UDLM)} & \multicolumn{2}{c}{CelebA-64 (DDPM)} \\
 & time & mem & time & mem & time & mem \\
\midrule
AdamW & 1.00 & 1.00 & 1.00 & 1.00 & 1.00 & 1.00 \\
Lion & 1.00 & 0.99 & 0.99 & 0.99 & \cellcolor[HTML]{FCF1F0}1.05 & 0.98 \\
Muon & \cellcolor[HTML]{FAE7E5}1.14 & 0.99 & \cellcolor[HTML]{FBECEB}1.08 & 0.99 & \cellcolor[HTML]{F2BDB9}2.04 & 0.98 \\
Schedule-Free & 1.00 & 1.00 & 1.01 & 1.00 & 1.01 & 1.00 \\
SOAP & \cellcolor[HTML]{F8DBD8}1.31 & \cellcolor[HTML]{FDF4F3}1.03 & \cellcolor[HTML]{F6D0CD}1.52 & \cellcolor[HTML]{FBEAE9}1.10 & \cellcolor[HTML]{EEA6A0}2.90 & 1.02 \\
MARS & \cellcolor[HTML]{FDF3F2}1.04 & 1.01 & 1.01 & 1.01 & \cellcolor[HTML]{FBE8E6}1.13 & 1.02 \\
MARS-M & \cellcolor[HTML]{FAE6E4}1.15 & 1.00 & \cellcolor[HTML]{FBEBEA}1.09 & 1.00 & \cellcolor[HTML]{F2BDB9}2.04 & 1.00 \\
\bottomrule
\end{tabular}

\end{table}

Per-step time is the median over all iterations past 500 of every seed, so
compilation and warmup are excluded; memory is the peak allocation over the
same range. Both are ratios against AdamW on the same task and hardware, and
LM1B is omitted because its runs used different hardware.

\section{Samples}
\label{app:samples}

Qualitative samples from the benchmarked models, one fixed sampling seed per
task so the outputs are comparable across optimizers. Weights follow each
task's evaluation protocol: EMA for CelebA-64 and raw final weights for the
text and molecule tasks. Because every optimizer decodes from the same noise, well-converged
models often emit an identical first sample.

\begin{figure}[htbp]
\caption{CelebA-64 faces, DDIM-100 as in the headline FID protocol, from the
same four initial noise draws for every optimizer, one column per optimizer.
DDIM is deterministic, so each column is the image that optimizer's model
assigns to the shared latents; the near-identical faces across optimizers show
that the models learned nearly the same latent-to-image map.}
\label{fig:celeba-samples}
\centering
\includegraphics[width=\textwidth]{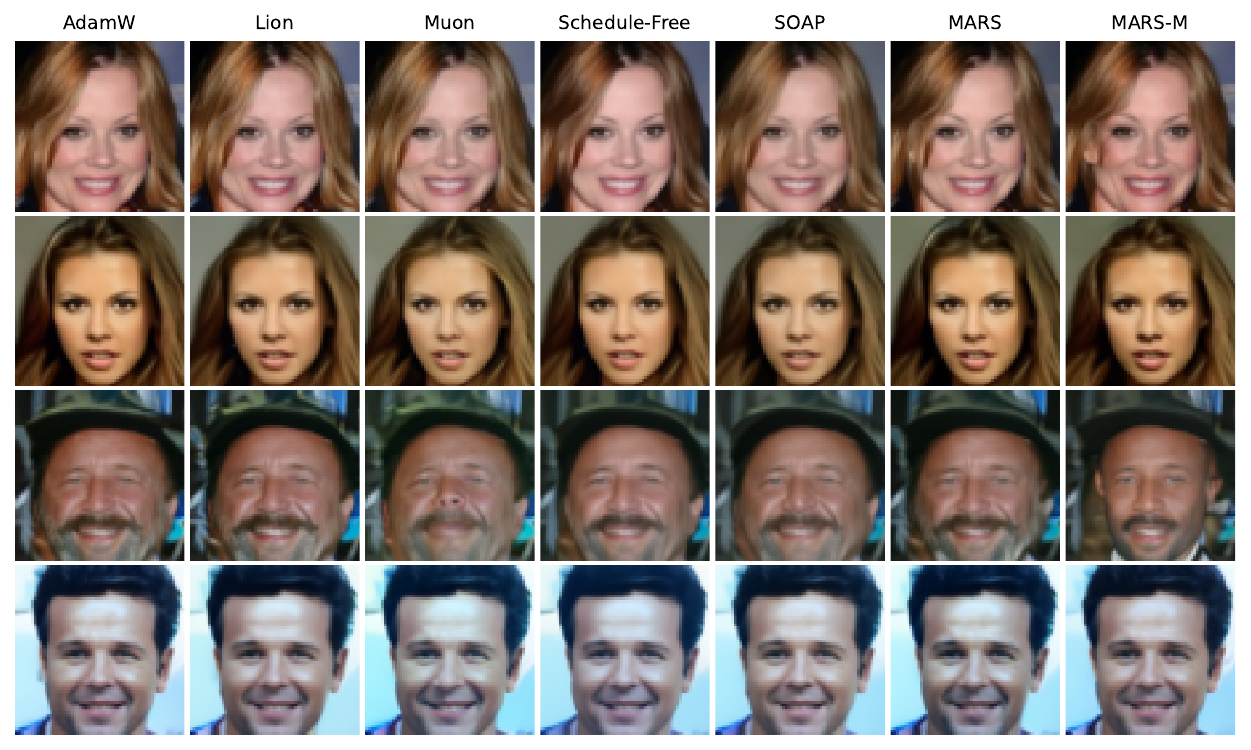}
\end{figure}

\begin{table}[htbp]
\caption{Unconditional text8 samples from each optimizer's model, 256 sampling
steps.}
\label{tab:text8-samples}
\centering
\small
\begin{tabular}{ll}
\toprule
optimizer & sample (first 64 of 256 characters) \\
\midrule
AdamW & \texttt{x iat sedokungu wolozzuy portiuoudeyoimoo qamigu kamea nlako xan} \\
Lion & \texttt{x ikprszdokungu woeozzly lortipoudeymimooljtmigubkamea nlako xao} \\
Muon & \texttt{ddicture only a low low short of a common pamigabiam a n and x o} \\
Schedule-Free & \texttt{x ist sedobunin woeozzly corti ouceymimoo jamigu kamea stano xao} \\
SOAP & \texttt{xxiatriedokunin colozzen port piusecormoo quilgaliam a alkelax o} \\
MARS & \texttt{x iap szdokungu joeozzly korti oudeymioxolqtmigubkamea nlako xao} \\
MARS-M & \texttt{x iakrszukkuningiourzzukklortiuiuceymiddoljamigulkatea nlako aan} \\
\bottomrule
\end{tabular}

\end{table}

\begin{table}[htbp]
\caption{Unconditional QM9 molecules from each optimizer's model, with the
fraction of its 1024 samples (as in Table~\ref{tab:qm9-quality}) that parse
as valid SMILES.}
\label{tab:qm9-samples}
\centering
\small
\begin{tabular}{lllr}
\toprule
optimizer & \multicolumn{2}{l}{samples (SMILES)} & valid \\
\midrule
AdamW & \texttt{C\#CC(=O)C1NC1C\#N} & \texttt{CC1CCC(=O)C(C)O1} & 100\% \\
Lion & \texttt{C\#CC(CO)C(=O)C\#N} & \texttt{CC1CCC(=O)C(C)O1} & 100\% \\
Muon & \texttt{C\#CC(CO)C(=O)C\#N} & \texttt{CC1CCC(=O)C(C)O1} & 99\% \\
Schedule-Free & \texttt{C\#CC(=O)C1NC1C\#N} & \texttt{CC1CCC(=O)C(C)O1} & 98\% \\
SOAP & \texttt{C\#CC(=O)C1NC1C\#N} & \texttt{CC1CCC(=O)C(C)O1} & 99\% \\
MARS & \texttt{C\#CC(=O)C1NC1C\#N} & \texttt{CC1CCC(=O)C(C)O1} & 99\% \\
MARS-M & \texttt{C\#CC(=O)C1NC1C=O} & \texttt{CC1CCC(=O)C(C)O1} & 99\% \\
\bottomrule
\end{tabular}

\end{table}

\begin{table}[htbp]
\caption{QM9 sample quality over 1024 molecules per optimizer (one
seed's weights, matching the evaluation protocol, one fixed sampling seed),
discussed in the QM9 results. Valid is the fraction that parses under RDKit
\citep{rdkit}, unique the distinct fraction among the valid, novel the
fraction of valid molecules absent from the training set, and FCD the Frechet
ChemNet Distance \citep{preuer2018fcd} to 10k training-set molecules.}
\label{tab:qm9-quality}
\centering
\small
\begin{tabular}{lrrrr}
\toprule
optimizer & valid $\uparrow$ & unique $\uparrow$ & novel $\uparrow$ & FCD $\downarrow$ \\
\midrule
AdamW & 99.6\% & 99.6\% & 8.6\% & 1.60 \\
Lion & 99.6\% & 99.5\% & 7.5\% & 1.69 \\
Muon & 99.3\% & 99.5\% & 8.8\% & 1.69 \\
Schedule-Free & 98.1\% & 99.6\% & 12.9\% & 1.47 \\
SOAP & 99.1\% & 99.3\% & 9.1\% & 1.66 \\
MARS & 99.3\% & 99.9\% & 7.0\% & 1.72 \\
MARS-M & 99.4\% & 99.7\% & 9.1\% & 1.54 \\
\bottomrule
\end{tabular}

\end{table}

\begin{table}[htbp]
\caption{LM1B samples from each optimizer's model (raw final weights of one
seed), length-conditioned to the median validation sentence length and
decoded with 1000 ancestral steps under the uniform-state reverse process.
The Lion model sampled here is one of the two seeds that diverge after the
curriculum switch, so its row samples the failed model.}
\label{tab:lm1b-samples}
\centering
\small
\begin{tabular}{ll}
\toprule
optimizer & sample (first 64 characters, median-length sentence) \\
\midrule
AdamW & \texttt{you shouldn ' t eat water or bangladeshi bins if they want to ge} \\
Lion & \texttt{film the financialonero war or. corporations proposition she the} \\
Muon & \texttt{you don ' t get water or fast, or if they want to study it and e} \\
Schedule-Free & \texttt{bharatiya pcs mohammed toro brothers or wedding corporations pro} \\
SOAP & \texttt{sbelle ' blows shown brothers or wedding corporations as she pic} \\
MARS & \texttt{you shouldn ' t imagine naming or amiss proposition if they want} \\
MARS-M & \texttt{she said she was metaphorical or technically corporationsy shevv} \\
\bottomrule
\end{tabular}

\end{table}

\section{Reproduction}
\label{app:repro}

The code is released at \codeurl: the four diffusion schemes, the seven
optimizers, the tuning and benchmark drivers, and the scripts that regenerate
every table and figure in this paper from the run logs. The README gives the
commands.

\paragraph{Assets and licenses.} Datasets: text8 \citep{mahoney2011text8} is
derived from a Wikipedia dump (text under CC BY-SA); QM9
\citep{ramakrishnan2014qm9} and LM1B \citep{chelba2013lm1b} are used as
distributed by their authors through the public Hugging Face mirrors named
in Appendix~\ref{app:data}; CelebA \citep{liu2015celeba} is used under its
non-commercial research agreement. Software: PyTorch and RDKit
(BSD-3-Clause); \texttt{diffusers}, \texttt{transformers},
\texttt{datasets}, \texttt{torch-fidelity}, \texttt{pytorch\_optimizer} and
\texttt{schedulefree} (Apache-2.0); Optuna, \texttt{lion-pytorch} and
\texttt{fcd\_torch} (MIT). Versions are pinned in the released code.

\end{document}